\documentclass[twocolumn]{article}

\usepackage{comment}
\usepackage{hyperref}
\usepackage{amssymb}
\usepackage{amsmath,amsfonts}
\usepackage{algorithmic}
\usepackage{graphicx}
\usepackage{textcomp}
\usepackage{algorithm}
\usepackage{algorithmic}
\usepackage{soul}
\usepackage[graphicx]{realboxes}
\usepackage{rotating}
\usepackage{subfigure}
\usepackage{caption}
\usepackage{svg}
\usepackage[normalem]{ulem}
\usepackage{appendix}
\usepackage{authblk}
\usepackage{amssymb}
\usepackage{latexsym}
\usepackage{url}
\usepackage{xcolor}
\usepackage{multirow}

\begin{document}

\title{Model-based inexact graph matching on top of CNNs for semantic scene understanding}

\author[1]{Jeremy Chopin}
\author[1]{Jean-Baptiste Fasquel}
\author[2]{Harold Mouchère}
\author[3]{Rozenn Dahyot}
\author[4]{Isabelle Bloch}

\affil[1]{Université d'Angers, LARIS, SFR MATHSTIC, F-49000 Angers, France}
\affil[2]{Nantes Université, École Centrale Nantes, CNRS, LS2N, UMR 6004, F-44000 Nantes, France}
\affil[3]{Department Computer Science, Maynooth University, Ireland}
\affil[4]{Sorbonne Universit\'e, CNRS, LIP6, Paris, France}

\date{}

\maketitle

\begin{abstract}
Deep learning based pipelines for semantic segmentation often ignore structural information available on annotated images used for training. 
We propose a novel post-processing  module enforcing structural knowledge about the objects of interest to improve segmentation results provided by deep neural networks (DNNs). This module  corresponds to a ``many-to-one-or-none'' inexact graph matching approach, and is formulated as a quadratic assignment problem. 
Our approach is compared to a DNN-based segmentation on two public datasets, one for face segmentation from 2D RGB images (FASSEG),  and the other for brain segmentation from 3D MRIs (IBSR).
Evaluations are  performed using two types of structural information: distances and directional relations that are user defined, this choice being a hyper-parameter of our proposed generic framework.   
On FASSEG data, 
results show that our module improves accuracy of the DNN by about 6.3\% i.e. the Hausdorff distance (HD) decreases from $22.11$ to $20.71$ on average.
With IBSR data, the improvement is of 51\% better accuracy with HD decreasing from $11.01$ to $5.4$.
Finally, our approach is shown to be resilient to small training datasets that often limit the performance of deep learning methods: the improvement increases as the size of the training dataset decreases.

\end{abstract}

\section{Introduction}
\label{intro}

Semantic segmentation is a task in computer vision where an image is divided into meaningful objects for image analysis and visual understanding. In the recent years, deep neural networks (DNN), more specially convolutional neural networks (CNN), have been widely used in  computer vision \cite{GoodBengCour16} and provided remarkable improvements for the task of semantic segmentation \cite{GARCIAGARCIA201841} compared to traditional methods such as random forest \cite{MO2022}.
Such DNN are trained in an end-to-end procedure and many different architectures have been proposed to be applied on a wide range of applications, such as medical imaging, autonomous driving or face labelling. 
As an example, Ronneberger et al. \cite{unet2015} have proposed the U-Net architecture for the semantic segmentation of biomedical images, which used context and spatial information to produce, to some extent, qualitative segmentation when few data are available.
Through a set of convolution layers, semantic segmentation with Convolutional Neural Networks (CNNs) is intrinsically based on information 
embedded at low-level, \textit{i.e.} at pixel and its neighborhood levels. 
CNNs do not explicitly model the structural information available at a higher semantic level, for instance the relationships between annotated regions that are present in the training dataset.
More recently Vision Transformers (VTs)  have emerged  as an alternative to CNNs 
for semantic image segmentation \cite{SegmenterICCV2021,2021_segFormer,2022_Unetr}.

High-level structural information may include spatial relationships between different regions (e.g. distances, relative directional position)~\cite{bloch2015} or relationships between their properties (e.g. relative brightness, difference of colorimetry)~\cite{fasquel2018,fasquel2017}.
This type of high-level structural information is very promising~\cite{DERUYVER2009876,NEMPONT20131, bloch2015,fasquel2018,fasquel2017}  and it has  found applications in medical image understanding~\cite{Colliot2006,fasquel2006,Moreno2008} but also in document analysis (e.g. \cite{delaye2011, julcaaguilar2020} for handwriting recognition) or in scene understanding (e.g. for robotics \cite{kunze2014combining}). 
In some domains, the relations between objects have to be identified to recognize the image content~\cite{julcaaguilar2020} but in other domains these relations help the recognition of a global scene as a complementary knowledge~\cite{delaye2011, kunze2014combining, fasquel2018, fasquel2017}. Our work falls in this second category.

This high-level information is commonly represented using graphs, where vertices correspond to regions, and edges carry the structural information. A challenge is therefore to compare relationships produced by a DNN  (CNN or VT) performing semantic segmentation  with the relationships observed in the training dataset, in order to identify more accurately regions of interest. The semantic segmentation problem turns then into a region or node labeling problem, often formulated as a graph matching problem \cite{lezoray2012, fasquel2018, fasquel2017}.

In this paper, we propose a new approach involving a graph-matching-based semantic segmentation applied to the probability map produced by DNNs for semantic segmentation. Our motivation is to explicitly take into account this high-level structural information observed in the training dataset. 
High-level structural information is only partially captured with convolutional layers (i.e. depending on the size of the considered receptive field). On the other hand, positional encoding in VTs captures better structural information but at a cost of an increase in computational complexity, requiring large training sets.
Our proposal aims at improving the semantic segmentation of images, in particular when the size of the training dataset is low.
As such, our work also addresses, to some extent, one key limitation of deep learning: the requirement of a large and representative dataset for training purposes, this being often addressed by generating more training data using data augmentation~\cite{surveydataaug19} or by considering a transfer learning technique \cite{TransferLearningGeneral}. %In this way,

By focusing on the high level global structure of a scene, our approach is expected to be less sensitive to the lack of diversity and 
representativity of the training dataset. This comes together with the size and the complexity of previously mentioned neural-network-architectures aiming at implicitly integrating such high-level information \cite{2021_segFormer,2022_convnet}: another motivation of this work is to contribute to solve this issue by explicitly integrating high-level structural information.
This paper extends preliminary works \cite{IPTA2020,ICPRAI2022Brain}, and our main contributions are: 
\begin{itemize}
\item Our method combines the high level structural information observed in the training dataset with the output of the semantic segmentation produced by a deep neural network. It uses a graph matching approach formulated as a quadratic assignement problem (QAP)~\cite{QAPref1,GraphQAP2016,QAP_gm}.
\item We deploy  two types of relationships for capturing structural information, illustrating how flexible and generic our approach is. This information is coupled, in a generic way, with heterogeneous region properties, 
not included in our preliminary works \cite{IPTA2020,ICPRAI2022Brain}.
\item Our method is shown experimentally to perform well on two  applications, for segmenting  2D (image data) and 3D data (volumetric data). Compared to our preliminary works \cite{IPTA2020,ICPRAI2022Brain}, our method is coupled with various recent DNN-based semantic segmentation networks, to show its relevance in various contexts. The baselines DNNs tested  with our novel module include architectures designed to deal with small datasets such as the CNN U-Net \cite{unet2015} and the enhanced  U-Net+CRF \cite{CRF2015},  as well as the VT architecture UNETR \cite{2022_Unetr}.
In our experiments, both CNNs and VTs are shown experimentally to have improved accuracy when used in combination with our proposed novel post-processing module.
\item We assess the robustness of our approach with decreasing sample size of the training dataset that affects the performance of the DNN  used as input to our module.
\item The open-source code and data  are shared with the community\footnote{\url{https://github.com/Jeremy-Chopin/APACoSI/}}.
\end{itemize}
Section \ref{sec:related} provides an overview of related works. 
The proposed method is then detailed in Section \ref{sec:method}. Experiments are presented in Section \ref{sec:experience}. Section \ref{sec:discussion} provides a discussion about this work before concluding.

\section{Related works}
\label{sec:related}

The task of semantic segmentation is an active field of research with many applications and new methods are often proposed.
As an example, for the task of semantic segmentation of subcortical structures of the brain, DNN models have been proposed  and led to very promising results.
Several architectures have been used, but often require a large quantity of memory which can be a limitation in the case of large images (3D MRI) \cite{YEE2022}. 
In order to overcome this limitation, the segmentation is applied on 2D slices or 3D patches of the image which are then fused to create the final segmentation.
However, this approach presents some drawbacks as patches only contain local information and lack spatial context \cite{YEE2022}. 
This is why recent work as in \cite{YEE2022} propose to reduce the memory requirements by detecting bounding boxes for both hemispheres of the brain, then segment one hemisphere and use this segmentation to segment the other hemisphere by reflecting it thought the sagittal plane. 
Others recent works aim at using the attention mechanism and multi-scale feature maps with Transformer and CNN to obtain more precise segmentation as in  \cite{LI2021, 2022_Unetr}.

For other tasks such as face labelling, semantic segmentation can be, to some extent, challenging when boundaries between regions of interests are ambiguous, when there are complex structures or a problem of class imbalance between labels. Recent works such as \cite{Shiping2021} propose, for the task of face labelling, to use sub-networks applied to different parts of the face so that they can learn rich features and also reduce competition between similar elements of the face (e.g. right eye and left eye).

Many techniques used for segmentation suffer from over-segmentation where regions of interest are splitted into sub-regions, and potential ambiguities when semantically different objects have similar shape and appearance. To cope with such problems, structural information can be useful. In this context, most related works based on structural relationships consider first non semantic oversegmentations, using algorithms such as meanshift, SLIC or watersheds \cite{DERUYVER2009876,Bloch2012,fasquel2017}.
Then, from this oversegmentation,  relationships between regions are built and then compared, together with region properties (e.g. mean intensity), to the ones of a model in order to identify each region. Such oversegmentations do not provide any semantic information in contrast to neural networks dedicated to semantic segmentation that provide a probability vector per pixel~\cite{unet2015}. To our knowledge, close works to ours integrating explicitly spatial relationships at the output of such a semantic segmentation network propose the use of conditional random fields (CRFs)~\cite{LIU20152983,CRF2015}. CRFs are limited to local neighborhoods, acting like a spatial regularization technique applied on the neural network output. As previously mentioned, some other recent works aim at implicitly integrating spatial information within the neural network. We can mention the CPNet \cite{ContextPrior2020}, the PSPNet \cite{2017_PSPNet} and proposals regarding vision transformers \cite{NIPS2017_transformer,2021_segFormer, 2022_Unetr}. They are mainly based on the notion of attention map \cite{NIPS2017_transformer} or affinity map \cite{ContextPrior2020} to be learned, to guide the segmentation.

The main issue is the complexity \cite{2022_convnet} and the parameterization of the architecture, in order to guarantee that, depending on the application domain, spatial information is appropriately captured (e.g. large enough receptive field for capturing long range spatial relationships). Recent work as in \cite{ELJURDI2021} uses prior information (shape or spatial information) embedded in the loss function used for the training procedure to increase the accuracy of the final segmentation. By contrast, our approach considers high-level relationships between image regions, directly observed in the annotated dataset, independently from their nature and their range (e.g. distances between regions), and aims at applying them on the top of the semantic segmentation networks. 

Our approach aims at matching regions produced by a neural network for a query image, with the labeled regions in our annotated dataset, by integrating both spatial relationships between regions. A common situation is the over-segmentation of the query image providing more regions than expected in the  model, and this situation requires a ``many-to-one'' inexact graph matching strategy.

Graph matching has been widely studied~\cite{surveygraph2014}, in particular in computer vision for various applications such as shape analysis, 3D-recognition, video and image database indexing~\cite{lezoray2012,Iodice20151074,Madi2016, fasquel2017,IPTA2020}. 
Various approaches have been proposed to graph matching: some use the notion of graph edit distance~\cite{lezoray2012,SERRATOSA2015204} or EDA \cite{PR-02,RC:PR-05}, while others formulate graph matching as a constraint satisfaction problem \cite{DERUYVER2009876}
or a quadratic assignment problem (QAP)~\cite{QAPref1, GraphQAP2016,QAP_gm}. All these approaches are expressed with a cost function to be minimized and/or a set of constraints to be verified. In this paper, we formulate graph matching as a quadratic assignment problem, as stated in \cite{QAP_gm}. 

Some recent approaches solve graph matching with machine learning (e.g. graph neural networks \cite{BACCIU2020203}) where they achieve exciting performances by learning an efficient node representation as described by \cite{LAURA2022, LIU2023}. In such a case, a set of graphs is provided together with node labels for training purposes. Although promising for many application domains \cite{GOYAL201878,graphUnet2019,ijcai2019}, large and representative training datasets of annotated graphs are required. Another difficulty is the definition of the appropriate architecture (choice of the building blocks, including the selection of the neighborhood aggregation strategy and the number of layers), and the management of both node and edge information, while edge features (related to region relationships) are often ignored \cite{ijcai2019-410}. Some recent work, as the one by \cite{LIU2023}, aims at learning the graph structure by performing a graph learning operation but, to our knowledge, there is no structural analysis of the final matching and the structural consistency is not guaranteed.

A difficulty associated with  graph matching is its combinatorial optimization.
As considered by many related works \cite{NOMA20113, Bloch2012,Iodice20151074,fasquel2017}, we use a two step approach: (i) finding a first correspondence (``one-to-one'' matching),  (ii) which is then refined by matching remaining nodes (``many-to-one'' matching). 
In contrast to non semantic based oversegmentation that are often considered,  the semantic segmentation provided by  neural networks will help our approach to reduce the search space for finding the optimal solution.

\begin{figure*}[!ht]
    \centering
    \includegraphics[width=0.95\textwidth]{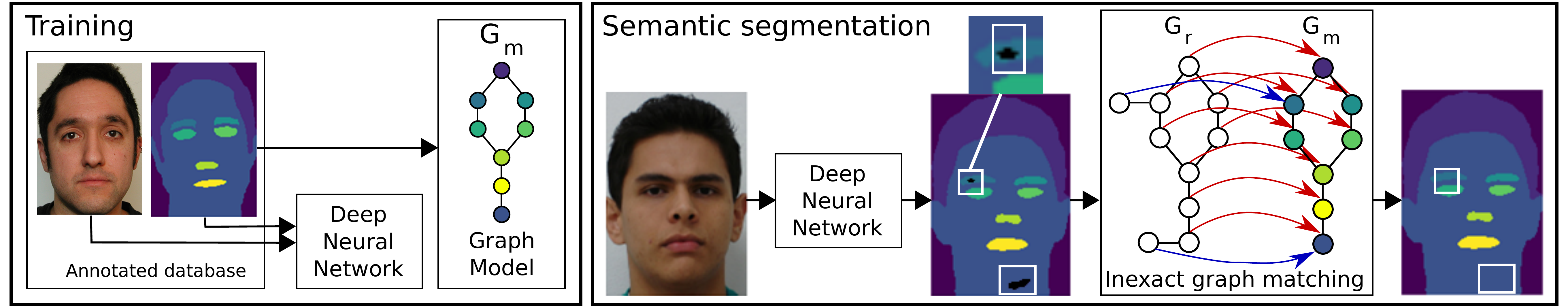}
    \caption{Overview of the proposed approach. Left: The training concerns both the deep neural network (training with both input image and expected output one) and the $G_m$ graph model (training with annotated images only). Right: Semantic segmentation combines the trained neural network and inexact graph matching (between graphs $G_r$ and $G_m$) for segmenting and identifying image regions.}
    \label{fig:overview}
\end{figure*}

\section{Proposed Method} 
\label{sec:method}

Figure \ref{fig:overview} provides an overview of our framework. 
The deep neural network is trained to perform semantic segmentation using an annotated dataset. Structural information, such as spatial relationships, is encoded in a graph model $G_m$ that captures the observed relationships between regions in the annotated training dataset. Vertices and edges  correspond respectively to regions of the annotated dataset and spatial relationships between them.

When performing the semantic analysis of an unknown image (cf. Figure \ref{fig:overview}-right), the neural network provides a segmentation proposal, from which a hypothesis graph $G_r$ is built. This hypothesis graph is then matched to the model graph. The purpose is to match the vertices (and thus the underlying regions) produced by the neural network with those of the model, to provide a relabelling of some of the regions (with many-to-one inexact graph matching). This matching, guided by both vertex and edge information, leads to a final semantic segmentation map  with high-level structural information as initially observed in the training dataset.

 Section \ref{subsec_dl} describes the output of the deep neural network which provides the initial semantic segmentation. We then focus on the construction of both hypothesis and model graphs, in Section \ref{subsec:graphhypo}. Finally, in Section \ref{subsec:matching}, we detail the proposed inexact-graph-matching step.

\subsection{Deep neural network}
\label{subsec_dl}
The deep neural network takes an input image or volume and produces, as output, a tensor $S\in \mathbb{R}^{P \times N}$ with $P$ the dimensions of the input ($P = I \times J$  pixels for 2D images, or $P = I \times J \times K $ voxels in 3D volumes) and $N$ is the total number of classes considered for segmentation.

At each pixel location $p$, the value $S(p,n) \in [0,1]$ is the probability of belonging to class $n$  with the constraints:
$$
\left( \forall n \in \{ 1,\ldots, N\} ,\  0\leq S(p,n)\leq 1 \right) \wedge  \left(  \sum_{n=1}^ N S(p,n) =1 \right)
$$

The segmentation map $\mathcal{L}^{*}$ selects the label $n$ of the class with the highest probability:

\begin{equation}
 \mathcal{L}^{*}(p)=\arg\max_{n \in \{1,\ldots, N\}} S(p,n)
\end{equation}
The deep neural network is trained using a dataset of examples containing both input images and their corresponding annotations (see Figure \ref{fig:overview}-left).
Note that this step can be achieved using any deep neural network dedicated to semantic segmentation, such as, for instance, U-Net \cite{unet2015}, nnU-Net \cite{nnUNET}, segNet \cite{segnet} or previously mentioned neural networks. 
%We have used U-Net in our experiments.

\subsection{Graph}
\label{subsec:graphhypo}

\begin{figure}[!ht]
    \centering
    \includegraphics[width=0.5\textwidth]{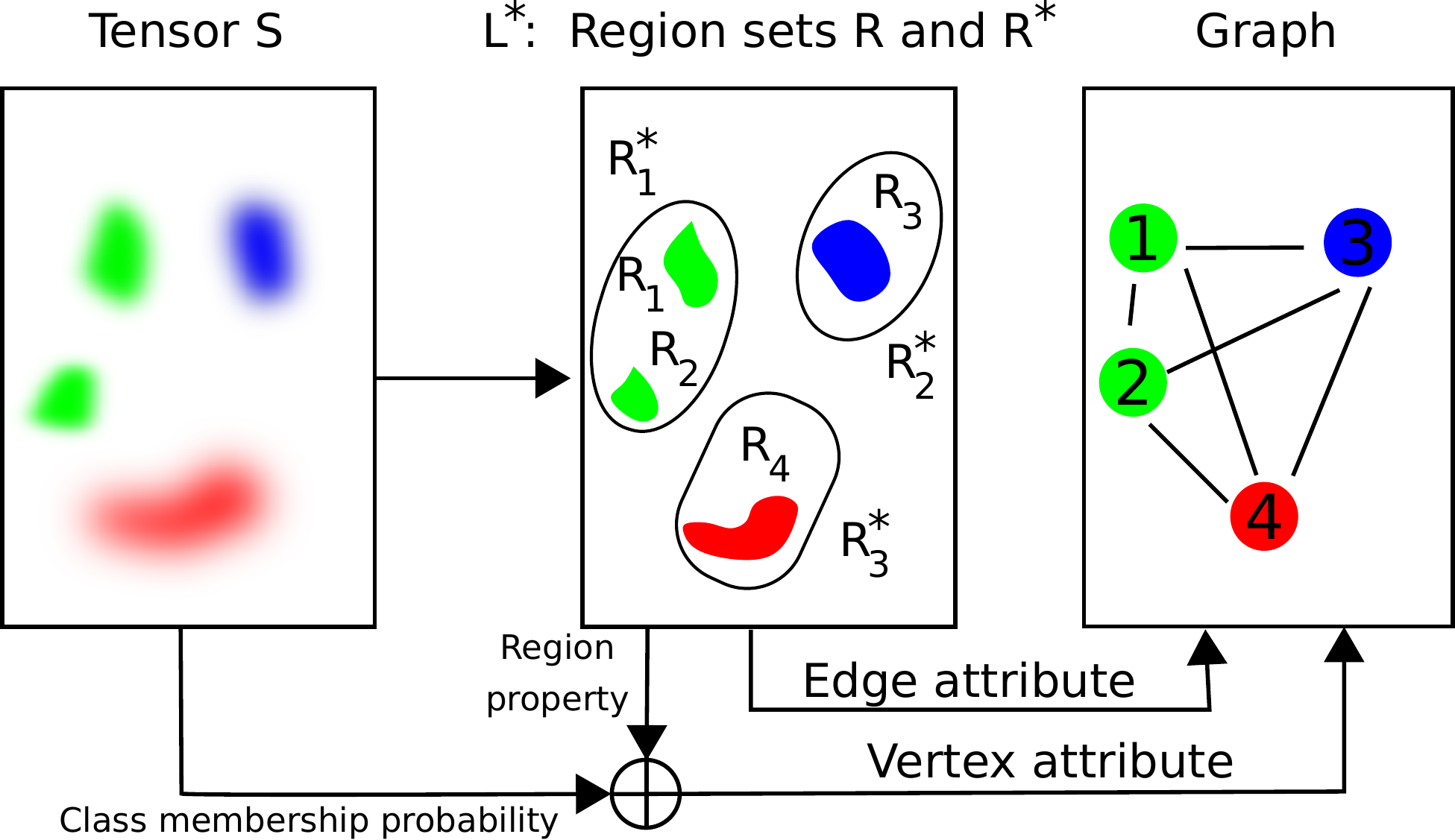}
    \caption{Graph construction ($G_r$) from the tensor S (output of the segmentation network) and the resulting $\mathcal{L}^{*}$ entity. Each point of the left image is associated to the probability vector represented by color intensity. $R^*_1$ is the set of regions (regions $R_1$ and $R_2$) that belong to class $1$ (according to probabilities). Edge attributes of the graph are computed from spatial relationships between regions $R_i$. Vertex attributes are the concatenation of a DNN-based information (class membership probability vectors) and other region properties (e.g. shape, size), computed over each region $R_i$.}
    \label{fig:graphc}
\end{figure}

From the segmentation map $\mathcal{L}^{*}$, we define a set $R$ of all resulting connected components (see an illustration in Figure~\ref{fig:graphc}, with $R=\lbrace R_1,\ldots,R_4 \rbrace$ and $N=3$). 

We also define a set $R^*=\lbrace R^*_1,\ldots,R^*_N \rbrace$, where, for each class $n \in \{1,\ldots,N\}$, $R^*_n$ is a set of regions corresponding to the connected components belonging to class $n$ according to the neural network (see Figure~\ref{fig:graphc}, where $R^*=\lbrace R^*_1, R^*_2, R^*_3 \rbrace$). 
This set $R^*$ is used for constraining the graph matching as described in Section \ref{sec:one2one}.

From the set $R$, a structural representation is built and modeled by the graph $G_r=(V_r,E_r,\alpha_v,\alpha_e)$, where $V_r$ is the set of vertices, $E_r$ is the set of edges, $\alpha_v$ is a function from $V_r$ into $A_v$ which associates to every vertex attribute values in $A_v$, and similarly $\alpha_e$ a function from $E_r$ into $A_e$ defining edge attributes. Each vertex $v\in V_r$ is associated with a region $R_v\in R$ with a normalized attribute, provided by the function $\alpha_v$ ($\forall v\in V_r,\ \alpha_v(v)\in A_v,\ A_v=[0,1]^{d_v}$, where $d_v$ is the number of attributes). As illustrated in our experiments, several properties can be considered, this being an hyperparameter of the proposal, to be adapted to the application. For instance, properties can be related to the output of the DNN (membership probability - see Figure \ref{fig:graphc}-left) or to the region shape and size for instance (see Figure \ref{fig:graphc}-middle). A key property is the average membership probability vector, defined by $S_{r,v}$. $S_{r,v}$ is a vector computed using both the region $R_v$ and the tensor $S$ (preliminary DNN-based segmentation, see Figure~\ref{fig:graphc}): 

\begin{equation}
\forall v \in V_r, \forall n \in \{1,\ldots, N\}, S_{r,v}[n] = \frac{1}{|R_v|} \sum_{p \in R_v} S(p,n)
\label{eq:classmember}
\end{equation}

The entity $S_{r,v}$ can be assimilated to a probability vector because $\sum_{i\in {1,\ldots, N}}S_{r,v}[i]=1$ and $\forall i \in \{1,\ldots, N\},\ 0\leq S_{r,v}[i]\leq 1$.

We consider a complete graph where each edge $e=(i,j) \in E_r$ has a normalized attribute defined by the function $\alpha_e$, associated with a relation between the regions $R_i$ and $R_j$ (cf. Figure~\ref{fig:graphc}), where $\forall e\in E_r,\ \alpha_e(e)\in A_e, A_e=[0,1]^{d_e}$, $d_e$ being the number of edge attributes. 

We define two possible functions $\alpha_e$  in our experiments, capturing the relative directional position or the trade-off between the minimal and maximal distances found between two regions.  The choice of the function $\alpha_e$ is an hyperparameter in our method that can be tuned to improve performance for the considered application (cf. Section~\ref{sec:experience}).

The model graph $G_m=(V_m,E_m,\alpha_v,\alpha_e)$ is composed of $N$ vertices (one vertex per class) and is constructed from the training set, with normalized vector attributes for vertices and edges, similarly to $G_r$. The attribute of a vertex is the average value of the properties computed over the training dataset. For the part dealing with the membership probability vector (from tensor $S_{m,v}$ related to the perfect segmentation provided by the annotations), it is also the average value, consisting of a vector of dimension $N$ with only one non-zero component (with value equal to $1$), associated with the index of the corresponding class. The edges are obtained by calculating the average relationships (in the training set) between the regions (according to the  relation $\alpha_e$ considered).
The main difference between $G_m$ and $G_r$ is that $G_m$ is derived from prior knowledge and fixed (for a given domain of application), while $G_r$ is derived from a single new instance to analyse and its prediction computed by the neural network. Therefore, $G_r$ may have more vertices than $G_m$  (oversegmentation), and is specific to each sample. Vertex and edge attributes are computed as for $G_m$.

\subsection{Matching with the model graph}
\label{subsec:matching}

The purpose of the graph matching operation is to label vertices, thus the underlying associated region, from the graph $G_r$ using the graph model $G_m$ which embeds the structural information considered (depending of the application), in order to produce a structurally consistent segmentation. Indeed, the most likely situation encountered is when more regions are found in the image associated with $G_r$  than in the model (i.e. $|V_r|\ge |V_m|$). To solve this, we propose here to extend the many-to-one inexact graph matching  strategy \cite{lezoray2012,IPTA2020} to a many-to-one-or-none matching.

The ``none'' term allows some vertices in $G_r$ to be matched with none of the vertices of the model graph $G_m$, which corresponds to removing  the underlying image region (e.g. merged with the background). Graph matching is here formulated as a quadratic assignment problem (QAP) \cite{GraphQAP2016}.

\begin{figure}[!ht]
    \centering
    \includegraphics[width=0.45\textwidth]{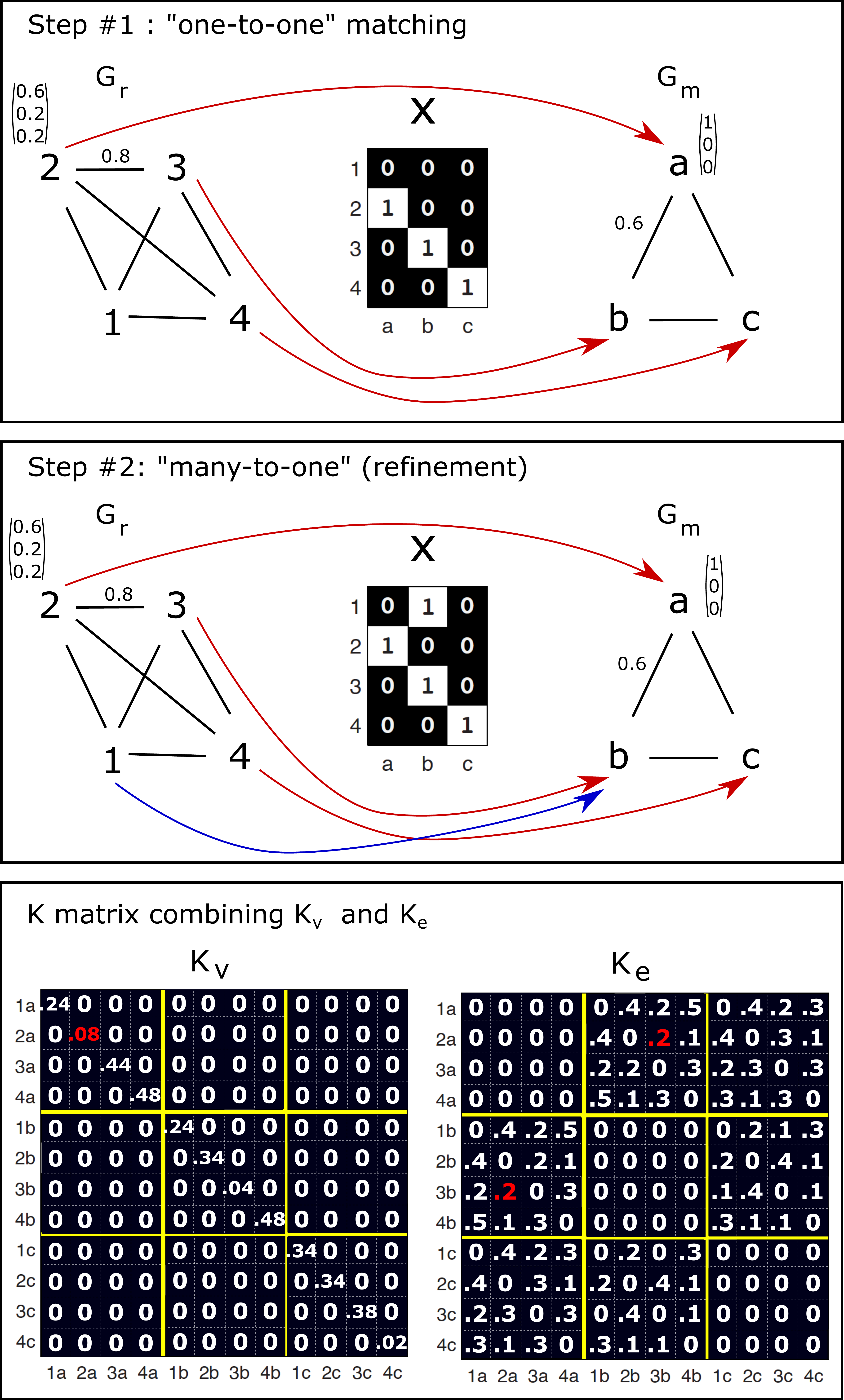}
    \caption{Graph matching formulated as a quadratic assignment problem (illustration inspired by~\cite{GraphQAP2016}). Our  proposal consists of two steps, where the $X$ matrix models a matching between $G_r$ and $G_m$ graphs. The first step (top) focuses on a one-to-one matching (each vertex of $V_m$ is associated with only one vertex of $V_r$). The second step (middle) aims at matching remaining vertices of $G_r$, leading to the final matching (many-to-one matching). For finding the optimal one-to-one matching (below), a $K$ matrix is used, combining both $K_v$ and $K_e$ matrices, which measure dissimilarities between vertices and edges, respectively. For the sake of clarity, only two edge attributes are reported.
  }
    \label{fig:gap}
\end{figure}

In our case, the concept of matching is represented by a matrix $X\in \{0,1\}^{|V_r|\times |V_m|}$, where $X_{ij}=1$ means that vertex $i\in V_r$ is matched with vertex $j\in V_m$. This is illustrated in Figure \ref{fig:gap} in two cases (``one-to-one'' and ``many-to-one'' matchings). 
The objective is to determine the best matching noted $X^*$ as follows:
\begin{equation}
X^*=\arg \min_{X} \left\lbrace\operatorname{vec}(X)^T K \operatorname{vec}(X) \right\rbrace
\label{eq:qap}
\end{equation}
where $\operatorname{vec}(X)$ is the column vector representation of $X$ and $T$ denotes the transposition operator.
This optimal matching is associated to the optimal matching cost $C^* = \operatorname{vec}(X^*)^T K \operatorname{vec}(X^*)$.

The matrix $K$ embeds the dissimilarity measures between the two graphs $G_r$ and $G_m$, at vertices (diagonal elements) and edges (non-diagonal elements): \begin{equation}
K=\lambda\ K_v+(1-\lambda)\ K_e
\label{eq:K}
\end{equation}
where $K_v$ embeds dissimilarities between vertex attributes produced by the function $\alpha_v$ (e.g. distance between class membership probability vectors, together with difference between some other region properties). More details for computing~$K$ can be found in~\cite{GraphQAP2016}. 
In the example shown in Figure \ref{fig:gap}, $K_v[1,1]=0.08$ (row and column named {\tt 2a} - mean square distance between $[0.6,0.2,0.2]$ and $[1,0,0]$) represents the dissimilarity
between vertex $2$ of $G_r$ and vertex $a$ of $G_m$, if one would match these two vertices. 
The matrix $K_e$ is related to dissimilarities between edges, and depends on the considered attribute function $\alpha_e$. For instance, in Figure~\ref{fig:gap}, $K_e[6,1]=0.2$ (row and column respectively named {\tt 3b} and {\tt 2a}) corresponds to the dissimilarity between edge $(2,3)\in E_r$ (scalar attribute whose value is 0.8) and edge $(a,b)\in E_m$ (scalar attribute whose value is $0.6$), if one would simultaneously match vertex $2$ with vertex $a$ and vertex $3$ with vertex $b$. In such a case, $ K_e[6,1]$ is computed using edge attributes:  $K_e[6,1]=|0.8-0.6|=0.2$. The $K_e$ terms are related to distances between regions (normalized in the final $K$ matrix).
The $\lambda$ parameter ($\lambda \in [0,1]$) allows weighting the relative contribution of vertex and edge dissimilarities (both $K_v$ and $K_e$ terms range between 0 and 1).

Due to the combinatorial nature of this optimization problem \cite{GraphQAP2016} (i.e. set of possible $X$ candidates in Equation \ref{eq:qap}), we propose a two-steps procedure, relying on the initial semantic segmentation provided by the neural network: 

\begin{enumerate}
    \item Search for an initial one-to-one matching (Figure \ref{fig:gap}-step 1).
    \item Refinement by matching remaining vertices, finally leading to a many-to-one-or-none matching (Figure \ref{fig:gap}-step 2).
\end{enumerate}

\subsubsection{Initial matching: one-to-one.}
\label{sec:one2one}

One searches for the optimal solution to Equation \ref{eq:qap} by imposing the following three constraints on $X$, thus reducing the search space for eligible candidates:
\begin{enumerate}
\item $\sum _{j=1}^{|V_m|} X_{ij} \leq 1$: some vertices $i$ of $G_r$ may not be matched.
\item $\sum _{i=1}^{|V_r|} X_{ij} = 1$: each vertex $j$ of $G_m$ must be matched with only one vertex of $G_r$.
\item $\lbrace X_{ij} = 1 \rbrace \Rightarrow \lbrace R_i\in R^*_j \rbrace$: vertex $i\in V_r$ can be matched with vertex $j\in V_m$ only when the associated $R_i$ region was initially considered by the neural network to most likely belong to class $j$ (i.e. $R_i\in R^*_j$). For instance, in the case of Figure \ref{fig:graphc}, only vertices related to regions $R_1$ and $R_2$ would be considered as candidates for class $1$ ($R^*_1$).
\end{enumerate}

The first two constraints ensure to search for a one-to-one matching, and thanks to the third constraint, one reduces the search space  by relying on the neural network: one assumes that it has correctly, at least to some extent, identified the target regions, even if artifacts may still have been produced as well (to be managed by refining the matching). This step allows us to retrieve the general structure of the regions (thus verifying the prior structure modeled by $G_m$) with a cost $C^{I}=\operatorname{vec}(X^I)^T K \operatorname{vec}(X^I)$ related to the optimal initial matching $X^I$ ($I$ stands for ``initial'').

\subsubsection{Refinement: many-to-one-or-none}
\label{sec:refinement}

Unmatched nodes are integrated to the optimal matching $X^I$ or removed (i.e. assigned to a ``background'' or ``none'' node) through a refinement step leading to $X^*$ considered in Equation~\ref{eq:qap}. This many-to-one-or-none matching is performed through an iterative procedure over the set of unlabeled nodes $U=\{k\in V_r \mid \sum _{j=1}^{|V_m|} X^I_{kj} = 0\}$. For each node $k\in U$, one searches for the best assignment, among all possible ones, related to the set of already labeled nodes $L=\{k\in V_r \mid \sum _{j=1}^{|V_m|} X^I_{kj} = 1\}$.

\begin{figure}[!ht]
    \centering
    \includegraphics[width=0.5\textwidth]{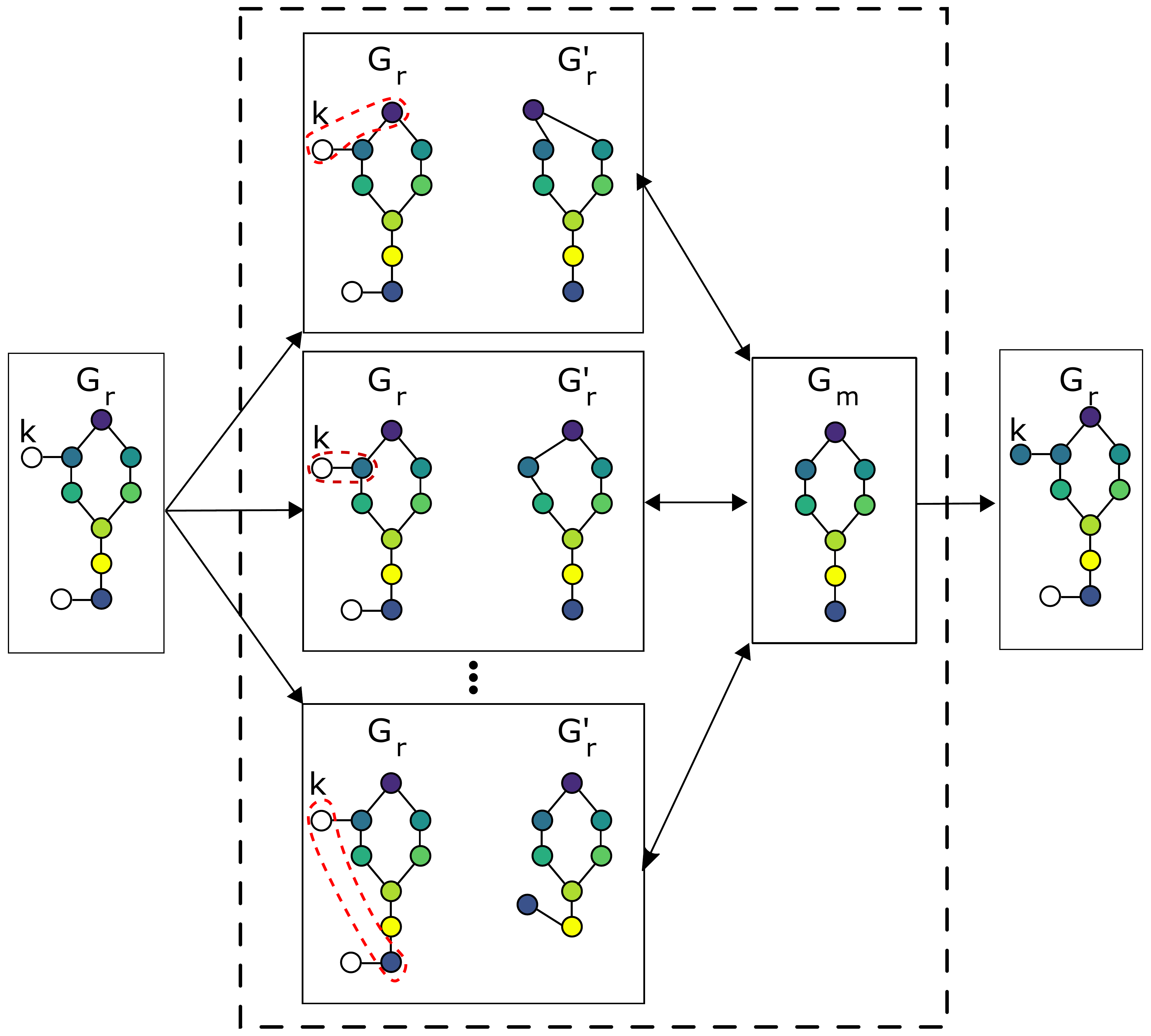}    \caption{Refinement: finding the best matching for a given unlabeled node $k\in U$ (white node). Only three possible matchings are reported for clarity (dashed surrounded nodes). The one in the middle is finally kept (smallest deformation of $G^{'}_r$ with respect to the model $G_m$).}
    \label{fig:overview_ref}
\end{figure}

Mathematically, the best label candidate for a given node $k\in U$ is:
\begin{equation}
  l^{*}_k=\arg \min_{l\in L}\left \lbrace  \operatorname{vec}(X^I)^T K_{k\rightarrow l} \operatorname{vec}(X^I) \right\rbrace
\end{equation}
where $K_{k\rightarrow l}$ corresponds to the matrix $K$ after having merged both underlying regions (i.e. $R_l = R_l\cup R_k$) and updated relations (leading to the graph $G^{'}_{r}$, where both $k$ and $l$ vertices are merged). The related cost associated to the merging of $k$ to $l^*_k$ is $C_{k\rightarrow l^*_k}$. Figure~\ref{fig:overview_ref} illustrates this iterative procedure.

The best candidate is retained if the related cost is smaller than a chosen threshold $T$, otherwise the related node $k$ is discarded (i.e. $k\rightarrow \emptyset$, $\emptyset$ corresponding to the ``none'' vertex, meaning that the underlying image region is merged with the background). The optimal matching is updated according to the condition:
\begin{equation}
%$$
X^*_{kl} = 
    \begin{cases}
        1 & \text{if}\ C_{k\rightarrow l^*_k} < T,\\
        0 & \text{otherwise}.
    \end{cases}
%$$
\label{eq:T}
\end{equation}
This enables to manage the removal of regions to be considered as artifacts for instance, which was not managed in our earlier work \cite{IPTA2020}.

Algorithm \ref{algo_qap} provides an implementation of the proposed refinement. For each unlabeled vertex $k \in U$, the optimal cost is initially set to infinity (Line 2). Then, for each candidate $l \in L$, one creates an image region (temporary variable $R'_l$) corresponding to the union of both unlabeled region and merging candidate region (Line 4). We update the dissimilarity matrix (leading to the temporary variable $K_{k \rightarrow l}$ - Line 5), and then compute the cost of this union (Line 6). If this union decreases the matching cost, the merging candidate is considered as the best one (Lines 8 and 9). After having evaluated the cost of the matching with the best candidate $l \in L$, we finally accept the resulting best matching, if the value of the associated cost is lower than the predefined threshold $T$ (Lines 12 to 16). If the cost is higher, the vertex $k\in U$ is discarded (and the image region is removed).

\begin{algorithm}[!ht]
\caption{Refinement algorithm}
\begin{algorithmic}[1]
\REQUIRE $U, L, T, X^I$
\FOR{$k \in U$}
\STATE $C^*_{k\rightarrow l} \leftarrow \infty$
\FOR{$l \in L$}
\STATE $R'_l \leftarrow R_l \cup R_k$
\STATE $K_{k \rightarrow l} \leftarrow \text{Update-K}(R'_l)$
\STATE $C_{k\rightarrow l} \leftarrow vec(X^I)^T \ K_{k \rightarrow l} \ vec(X^I)$
\IF{$C_{k\rightarrow l} < C^*_{k\rightarrow l}$}
\STATE $l^*_k \leftarrow l$
\STATE $C^*_{k\rightarrow l^*_k} \leftarrow C_{k\rightarrow l}$
\ENDIF
\ENDFOR
\IF{$C_{k\rightarrow l^*_k} < T$} 
\STATE $k \rightarrow l^*_k$ \COMMENT {$k$ is assigned}
\STATE $R_{l^*_k} \leftarrow R_{l^*_k} \cup R_k$
\ELSE
\STATE $k \rightarrow \emptyset$ \COMMENT {$k$ is discarded}
\ENDIF
\ENDFOR
\end{algorithmic}
\label{algo_qap}
\end{algorithm}

\section{Experiments}
\label{sec:experience}
Experiments focus on two applications and two types of structural information.  
Applications targeted are for the semantic segmentation of images of faces (FASSEG public dataset) and the segmentation of internal brain structures (IBSR public dataset), for 2D color images and 3D medical images, respectively.

Section \ref{subsec:proto} describes the evaluation protocol. Sections \ref{sec_fasseg} and \ref{sec_ibsr} detail experiments and results on the FASSEG dataset and the IBSR dataset, respectively. In each case, we first describe the dataset, then the structural information, the considered DNNs and finally the results.

\subsection{Protocol}
\label{subsec:proto}

In order to evaluate the performance of our approach, we compute the Dice score (DSC) and the Hausdorff distance (HD) for each class. The Dice score measures the similarity between the segmented region and the annotated one (value between 0 and 1 to maximize). The Hausdorff distance is a value to be
minimized (0 corresponding to a perfect segmentation). Compared to the Dice score, the HD emphasizes 
the case of a segmented region that would depict small connected components that are far away from the reference annotated region. One also compares results obtained by our approach with the ones obtained by considered DNNs.

We also study the behavior of our method with DNNs trained with fewer training samples. In order to do this, we consider different sizes of the training dataset for both experiments. We consider smaller training datasets, as reported in Table \ref{tab:datasetsize} (details regarding both datasets are given in Sections~\ref{sec_fasseg} and~\ref{sec_ibsr}).
\begin{table}[htb!]
\begin{center}
\small
\caption{
Dataset size, number of training images (validation) per training dataset size (A, B and C configurations), and number of images in the testing dataset.}
\label{tab:datasetsize}

\begin{tabular}{|c|c|c|c|c|c|}
\hline
 & dataset size & A & B & C & test\\
\hline
FASSEG &$70$ & 15 (5) & 10 (3)  & 5 (2) & 50 \\
\hline
IBSR & $18$ & 8 (4)  & 6 (3) & 4 (2) & 6 \\
\hline
\end{tabular}
\end{center}
\end{table}
For each training dataset size, the set of images used for testing purposes is always the same. For each configuration (A, B or C), results are averaged over 3 randomly selected training datasets, leading to 3 differently trained DNN-based neural networks. For instance, for the C configuration on IBSR, 3 random selections of 4 training images over the 12 available ones are considered (i.e. the initial dataset of 18 images without the 6 test images).

In our experiments, several DNN backbones are considered (hereafter detailed), selected among recent works with available implementations, either specialized for 2D color images or for 3D medical images.

\subsection{FASSEG}
\label{sec_fasseg}

\subsubsection{Data}
The FASSEG\footnote{ \url{https://github.com/massimomauro/FASSEG-dataset}.} public dataset focuses on the multi-class semantic segmentation of the face~\cite{fasseg2015-seg} as well as the estimation of its pose \cite{fasseg2017-pose}. In our study, we consider a subset of this dataset corresponding to a specific pose (the front view). 
This subset contains 70 images with manual semantic segmentation that, however, does not distinguish certain regions in annotated images (i.e. left eye and right eye, left eyebrow and right eyebrow). Therefore, we have refined the annotations in order to have a unique region per label, leading to the FASSEG-Instances public dataset (freely available\footnote{\url{https://github.com/Jeremy-Chopin/FASSEG-instances}}). For the sake of simplicity, the term FASSEG is used in
the rest of the paper. Eight classes are considered:
hair (Hr), face (Fc), left eyebrow (L-br), right eyebrow (R-br), left eye (L-eye), right eye (R-eye), nose and mouth.

\subsubsection{Structural information}
\label{sec:structFASSEG}
\begin{figure}[!htb]
\begin{center}
\begin{tabular}{cc}
\includegraphics[width=.45\linewidth]{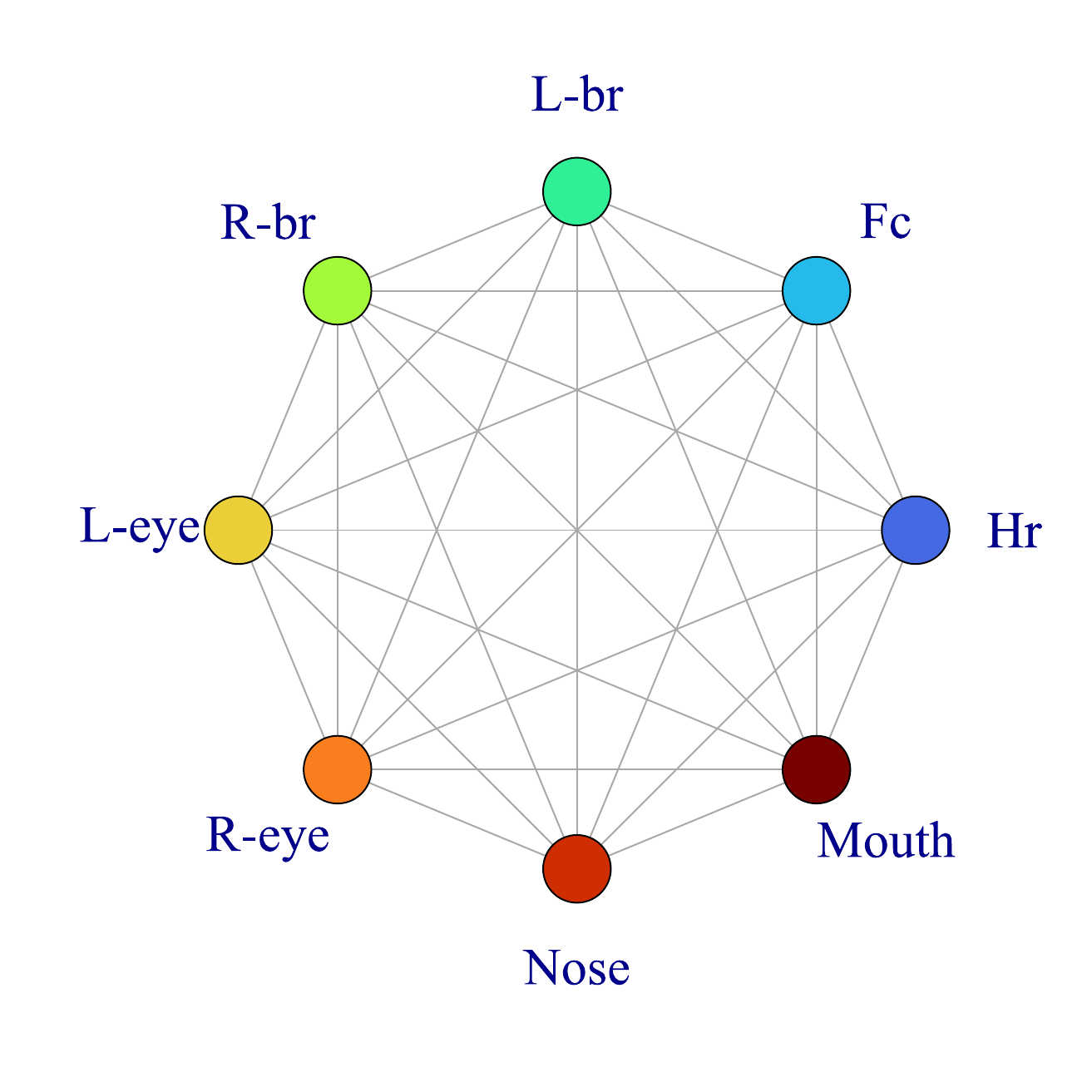}&
\includegraphics[width=.45\linewidth]{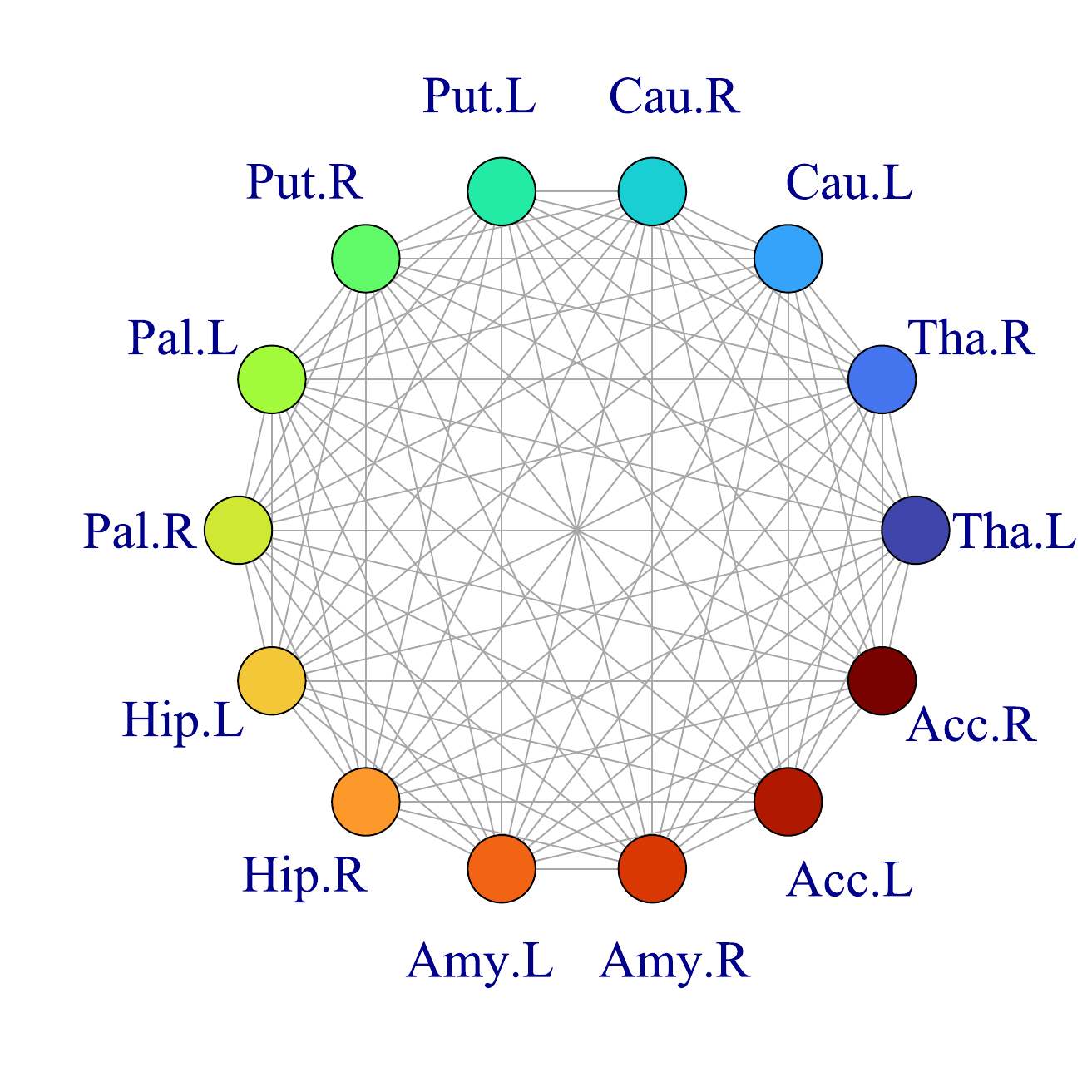} \\
FASSEG: $|V_m|=8$ & IBSR: $|V_m|=14$\\
\end{tabular}
\end{center}
\caption{The graph models $G_m=(V_m,E_m,\alpha_v,\alpha_e)$ are fully connected with $|E_m|=|V_m|^2$.  Note that the graph to be matched $G_r=(V_r,E_r,\alpha_v,\alpha_e)$ is likewise fully connected with $|E_r|=|V_r|^2$ and  $|V_r|\geq |V_m|$ (as detailed in experiments and reported in Tables \ref{tab:nb_vertices_fasseg} and \ref{tab:nb_vertices_ibsr}). 
}
\label{fig:Gms}
\end{figure}

While the structure of the scene is mainly carried by the edges and edge attributes, we encompass also vertex attributes in structural information. The model graph $G_m$ is a complete graph with $8$ nodes corresponding to the $8$ classes (see Figure \ref{fig:Gms}-FASSEG).

\begin{figure*}[!htb]
    \centering
    \includegraphics[width=\textwidth]{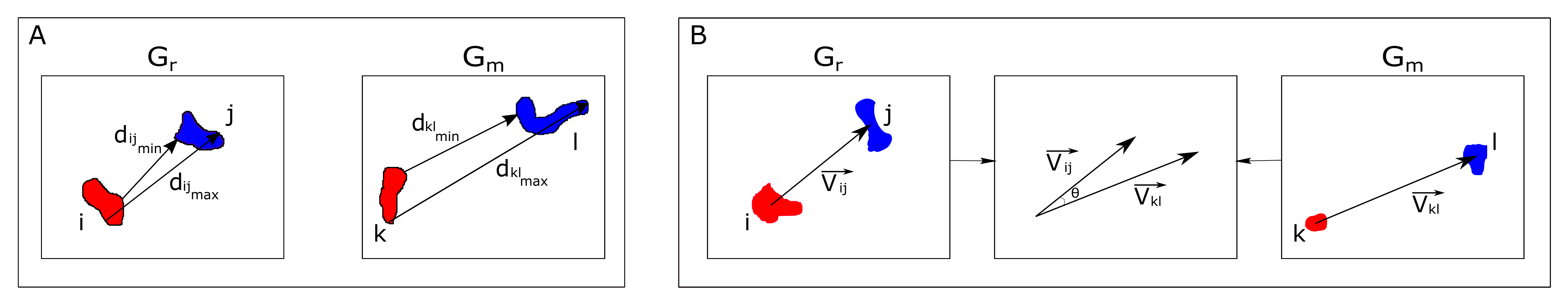}
    \caption{Spatial relationships considered in experiments. A: Relationship based on distances (corresponding to the ${}_eD_1$ dissimilarity function). B: Relationship based on relative directional positions (corresponding to the ${}_eD_2$ dissimilarity function).}
    \label{fig:distance_all}
\end{figure*}

Concerning vertices, the considered attribute is the class membership probability vector. The function $\alpha_v$ is defined by $\forall i \in V_r,\ \alpha_v(i)=S_{r,i}$ for the graph $G_r$ (defined in Equation \ref{eq:classmember}), and by $\forall i \in V_m,\ \alpha_v(i)=S_{m,i}$ for the graph $G_m$.
To compute $K_v$, we consider the dissimilarity function ${{}_vD_1}_{i}^{k}$ ($i\in V_r$ and $k \in V_m$):
\begin{equation}
\label{A_edt}
{{}_vD_1}_{i}^{k}=\frac{1}{N}\sum_{1\leq n\leq N}(S_{m,k}[n]-S_{r,i}[n])^2
\end{equation}
corresponding to the mean squared error (MSE) between the two class membership probability vectors.

The edge attribute, corresponding to a spatial relationship, involves two distances (leading to two components on an edge attribute), corresponding to the minimal and maximum distances between two regions $R_i$ and $R_j$ (see Figure \ref{fig:distance_all}-left):
\begin{equation}
    d_{min}^{(i, j)} = \frac{\min_{ p \in R_i, q \in R_j }(|p-q|)}{C_s}    \label{edt_min}
\end{equation}
\begin{equation}
    d_{max}^{(i, j)} = \frac{\max_{ p \in R_i, q \in R_j }(|p-q|)}{C_s}
    \label{edt_max}
\end{equation}
where $C_s$ corresponds to the largest distance observed in an image, ensuring that values range within $[0,1]$. As a result, $\forall e=(i,j)\in E, \alpha_e(e)=[d_{min}^{(i, j)},d_{max}^{(i, j)}]$.
Based on these relationships, the considered dissimilarity ${{}_eD_1}_{(i,j)}^{(k,l)}$ function is defined as:
\begin{equation}
\label{D_edt}
{{}_eD_1}^{(k,l)}_{(i,j)} = \lambda_e  \left( |d_{min}^{(i,j)} - d_{min}^{(k,l)}|\right) \ +(1- \lambda_e) (|d_{max}^{(i,j)} - d_{max}^{(k,l)}|)
\end{equation}
where $(k,l)\in V^2_m$, $(i,j)\in V^2_r$ and $\lambda _e$ is a parameter balancing the influence of the dissimilarities on both distances.

\subsubsection{Hyperparameters}

For the experiments on the FASSEG dataset, the values of $\lambda$ (Equation \ref{eq:K}) and $\lambda_e$ (Equation~\ref{D_edt}) are both set to 0.5 and 0.5.
The value of $T$ in Equation \ref{eq:T} is set to $1.01$. 
These parameters are chosen empirically and no optimization has been applied. 
The same parameter values have been used for all the DNN backbones without any specific tuning.

\subsubsection{DNN backbones}
For this dataset, we consider several DNNs, the code of which being available: U-Net \cite{unet2015}, U-Net combined with CRF as post-processing \cite{CRF2015}, PSPNet \cite{2017_PSPNet} and EfficientNet \cite{EfficientNet19}.

The U-Net is a well known and widely used architecture. For the experiments, a pytorch implementation was used~\cite{codeUnet_2019_3522306}, with 200 epochs for training the network and an early stopping strategy  applied to prevent over-fitting. The training process was stopped if there was no improvement on the loss (using Dice and cross-entropy loss functions) during 10 consecutive epochs. We also applied a median filter to remove tiny artifacts from the segmentation map provided by the U-Net.  Concerning the CRF-based post-processing \cite{CRF2015}, acting as a spatial regularization technique, a Gaussian filter of size 11 was applied and the CRF model was placed at the output of a pretrained U-Net and trained (for 50 epochs) to fine tune the whole model.

The PSPNet introduces a pyramid pooling module to capture information at different scales, corresponding to various ranges of spatial relationships between subregions (i.e. local and global context information). The final pixel classification task is then based on the fusion of the feature map and this multiscale information. The hyperparameters were those used in~\cite{2017_PSPNet} (use of a pretrained ResNet for the initial feature map computation, and a four-level pyramid pooling).

The EfficientNet is based on a compound scaling method allowing to appropriately  scale the network width, depth and resolution, in order to improve the accuracy of a DNN. In our experiments, we focused on the EfficientNet-B3 (pretrained on ImageNet) with hyperparameters set as in~\cite{EfficientNet19}, including a U-Net for decoding  (Eff-UNet~\cite{Eff-UNET20}).

The number of considered training, validation and test images is reported in Table~\ref{tab:datasetsize}, for the various training configurations.

\subsubsection{Results}
\label{results_fasseg}

\begin{table*}[!ht]
\caption{
Results on the FASSEG dataset, for different DNNs and training configurations. Results are given as Dice score (DSC) and Hausdorff distance (HD) with average values computed using results from 200 test images (the 50 test images with each DNN trained 4 times) and standard deviation in parentheses, together with minimum and maximum values (between brackets).}
\label{tab:FASSEG_results_globale}
\resizebox{\textwidth}{!}{%
\begin{tabular}{|c|c|c|c|c|c|c|c|c|c|c|c|c|}

\hline
Training & \multicolumn{4}{|c}{A training configuration} & \multicolumn{4}{|c}{B training configuration} & \multicolumn{4}{|c|}{C training configuration}\\

\hline
Method & \multicolumn{2}{|c}{DNN} & \multicolumn{2}{|c}{Proposal} & \multicolumn{2}{|c}{DNN} & \multicolumn{2}{|c}{Proposal} & \multicolumn{2}{|c}{DNN} & \multicolumn{2}{|c|}{Proposal}\\

\hline
Model & DSC$\uparrow$  & HD$\downarrow$  & DSC$\uparrow$  & HD$\downarrow$ & DSC$\uparrow$  & HD$\downarrow$ & DSC$\uparrow$  & HD$\downarrow$ & DSC$\uparrow$  & HD$\downarrow$ & DSC$\uparrow$  & HD$\downarrow$ \\

\hline

\multirow{2}{*}{U-Net}& 0.7 (0.05)& 37.4 (19.32)& 0.7 (0.05)& \textbf{35.79} (21.33)& 0.72 (0.06)& 43.02 (26.64)& \textbf{0.73} (0.05)& \textbf{32.59} (17.73)& 0.71 (0.07)& 54.13 (33.5)& \textbf{0.72} (0.07)& \textbf{37.55} (22.27)\\
& [0.55; 0.84] & [13.61; 118.11]& [0.55; 0.84] & [13.97; 124.19]& [0.58; 0.87] & [12.09; 141.68]& [0.6; 0.87] & [10.84; 121.02]& [0.48; 0.87] & [6.62; 162.95]& [0.48; 0.87] & [6.62; 135.96]\\

\hline

\multirow{2}{*}{U-Net + CRF}& 0.71 (0.05)& 35.4 (17.18)& 0.71 (0.05)& \textbf{34.07} (18.21)& 0.7 (0.05)& 39.87 (20.71)& \textbf{0.71} (0.05)& \textbf{34.48} (20.31)& 0.69 (0.06)& 60.01 (30.55)& 0.69 (0.08)& \textbf{43.99} (26.14)\\
& [0.57; 0.83] & [12.34; 93.76]& [0.58; 0.83] & [12.34; 122.05]& [0.56; 0.82] & [14.59; 115.12]& [0.57; 0.83] & [12.46; 118.16]& [0.54; 0.83] & [15.65; 157.32]& [0.13; 0.84] & [13.27; 182.14]\\

\hline

\multirow{2}{*}{PSPNet}& 0.83 (0.05)& 24.18 (16.58)& 0.83 (0.06)& \textbf{20.44} (14.82)& 0.82 (0.05)& 27.8 (23.0)& 0.82 (0.05)& \textbf{20.96} (15.63)& 0.76 (0.09)& 79.93 (47.03)& 0.76 (0.11)& \textbf{33.07} (25.32)\\
& [0.64; 0.98] & [2.4; 94.01]& [0.62; 0.98] & [2.4; 101.18]& [0.66; 0.98] & [2.78; 126.89]& [0.64; 0.98] & [2.78; 100.45]& [0.55; 0.98] & [3.05; 248.97]& [0.31; 0.98] & [3.05; 146.37]\\

\hline

\multirow{2}{*}{EfficientNet}& 0.84 (0.05)& 22.11 (17.0)& 0.84 (0.05)& \textbf{20.48} (17.04)& 0.83 (0.05)& 26.39 (18.51)& 0.83 (0.05)& \textbf{22.64} (18.82)& 0.81 (0.05)& 49.8 (33.88)& 0.81 (0.06)& \textbf{29.13} (23.27)\\
& [0.7; 0.98] & [2.38; 82.52]& [0.7; 0.98] & [2.38; 125.77]& [0.71; 0.98] & [1.6; 81.69]& [0.74; 0.98] & [1.6; 127.12]& [0.61; 0.99] & [1.36; 144.91]& [0.61; 0.99] & [1.36; 169.74]\\

\hline

\end{tabular}
}
\end{table*}

\begin{figure*}[!h]
    \centering
    \includegraphics[width=\textwidth]{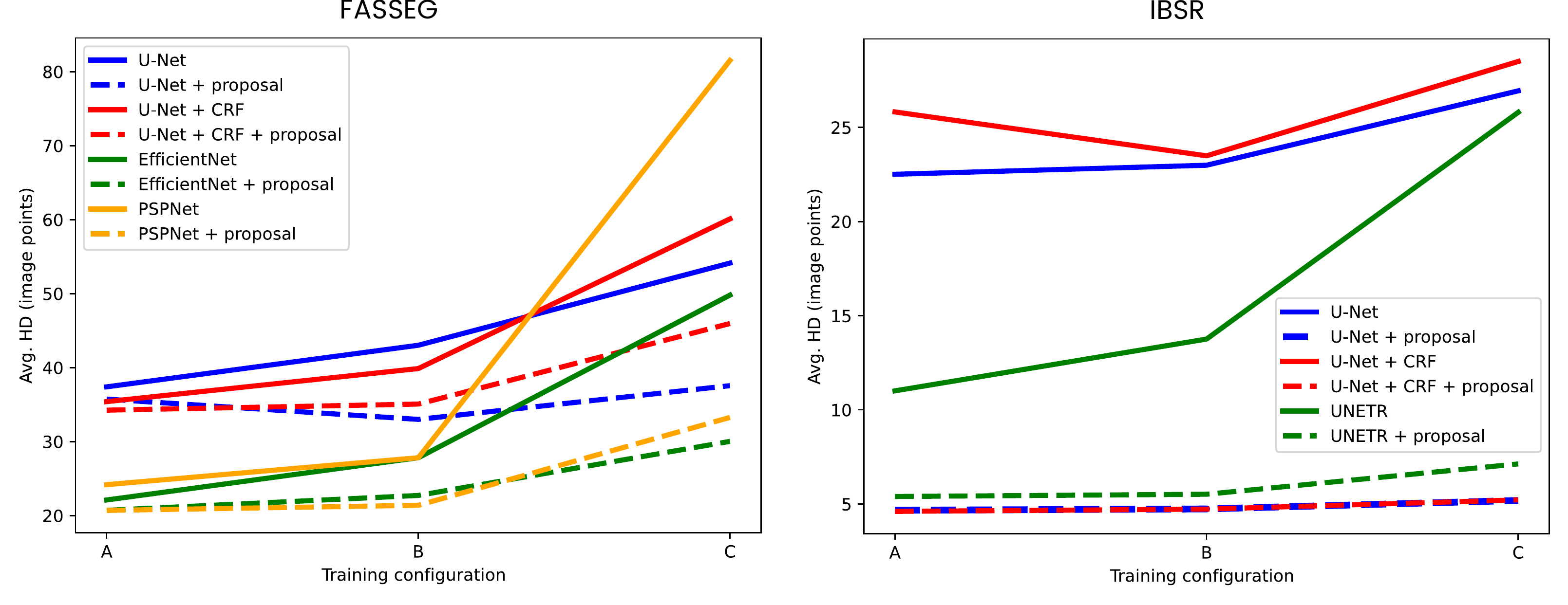}
    \caption{Average Hausdorff distance measured, for different training configuration, on the segmentation provided by DNNs and on the one resulting from our proposal (dashed lines) for both FASSEG and IBSR. When the training dataset is smaller, the improvement resulting from our proposal is higher.}
    \label{fig:hd_config_curves}
\end{figure*}

Table \ref{tab:FASSEG_results_globale} provides segmentation results for different DNN backbones, as well as for different sizes of the training dataset. 
The improvement appears significant in terms of Hausdorff distance, whatever the training configuration and the DNN, while it appears negligible in terms of Dice score. This means that segmentation errors consist of small regions that are relatively far away from the target region (hence significantly altering the Hausdorff distance). Note that, although recent DNNs such as EfficientNet and PSPNet lead to significantly better results than U-Net (both in terms of Dice score and Hausdorff distance), our proposal still improves the segmentation. 
The improvement increases with the reduction of the size of the training dataset (see the C configuration compared to the A one). As a result, for the most efficient DNN (EfficientNet), the Hausdorff distance decreases from $22.11$ to $20.71$ (reduction of $6.3$\%) for the A training configuration, from $26.39$ to $22.72$ (reduction of $13.8$\%) for the B one, and from $49.8$ to $30.03$ (reduction of $39.7$\%) for the C one.
This illustrates how well our approach  compensates for the lack of representativity when using a small training dataset: the smaller the training dataset is, the larger the improvement is. This trend is illustrated in Figure~\ref{fig:hd_config_curves}-FASSEG.

Table \ref{tab:FASSEG_parameters} reports the number of trainable parameters required for DNNs, to be compared with the small number of parameters used for building the model graph ($G_m$) from the annotated dataset (128 parameters corresponding to mean relationships between the 8 classes). Note that EfficientNet is a more complex architecture than the traditional U-Net, leading to a better Dice score at the cost of more trainable parameters (Table \ref{tab:FASSEG_results_globale}). PSPNet appears more efficient than U-Net and similar to EfficientNet, but with fewer parameters, underlying the importance of capturing different pieces of information at different scales. As previously underlined, whatever the DNN backbone, our proposal brings significant improvement in terms of HD, showing the importance of large scale information such as the high-level region relationships captured from annotated images. Table~\ref{tab:nb_vertices_fasseg} reports the number of vertices in $G_r$, observed in experiments, to be matched with those of the model graph $G_m$ ($|V_m|=8$ as illustrated by Figure~\ref{fig:Gms}). In most cases, this number increases as the training dataset size decreases because the DNN is less efficient, involving more region candidates (vertices) per class.

\begin{table}[!ht]
\begin{center}
\small
\caption{
Number of trainable parameters for each architecture used for the FASSEG dataset.}
\label{tab:FASSEG_parameters}

\begin{tabular}{c|c}

\hline
Architectures & Nb. trainable parameters \\
\hline
U-NET & 7 865 529\\
\hline
U-NET + CRF & 7 865 532\\
\hline
EfficientNet & 17 779 713\\
\hline
PSPNet & 3 849 420\\
\hline

\end{tabular}
\end{center}
\end{table}

\begin{table}[!ht]
\caption{
Average number of regions ($\overline{\#R}$) in the segmentation provided by the DNNs  considered for the FASSEG dataset (the ideal number of regions is 8) and the running time (Exec time in seconds) of our graph matching method to process these segmentation maps.
 The average number of regions and the average execution time ($\overline{\text{Exe}}$) considering each training configuration are reported with standard deviation (in parentheses) computed using results from 200 test images (the 50 test images with each DNN trained 4 times). Bold font indicates best results: as expected the DNN U-Net performs well with small training set (C training configuration). Note that execution time is proportional to the number of regions found.}
 \label{tab:nb_vertices_fasseg}
 \resizebox{\linewidth}{!}{  
\begin{tabular}{|c|c|c|c|c|}
\hline
Model & Metrics &  Configuration A & Configuration B & Configuration C\\
\hline

\multirow{2}{*}{U-Net (2015)} & $\overline{\#\text{R}}$ & 11.06 (2.56) & 12.12 (3.63) & \pmb{13.05} (4.79)  \\
 & $\overline{\text{Exe}}$ & 2.55 (1.77)  & 3.29 (2.52)  & \pmb{3.98} (3.44) \\

 \hline
 
\multirow{2}{*}{U-Net + CRF (2015)} & $\overline{\#\text{R}}$ & 11.18 (3.07) & 11.87 (3.69) & 14.01 (4.90) \\
 & $\overline{\text{Exe}}$ & 2.64 (2.13) & 3.12 (2.57) & 4.63 (3.50) \\

 \hline
 
\multirow{2}{*}{PSPNet (2017)} & $\overline{\#\text{R}}$ & 10.40 (3.14) & 10.75 (3.66) & 18.14 (7.41) \\
 & $\overline{\text{Exe}}$ & 2.20 (2.23) & 2.45 (2.62) & 8.02 (6.17) \\

 \hline
 
\multirow{2}{*}{EfficientNet (2019)} & $\overline{\#\text{R}}$ & \pmb{9.36} (1.95) & \pmb{10.41} (2.80) & 15.29 (8.03) \\
 & $\overline{\text{Exe}}$ & \pmb{1.45} (1.38) & \pmb{2.20} (2.00) & 6.01 (7.81) \\
 
\hline

\end{tabular}
}
\end{table}

Table \ref{tab:fasseg_ablation} reports the results we obtained for an ablation study where we compare the results of the segmentation after the initial "one-to-one-or-none" matching (Inter) and after the refinement operation leading to a "many-to-one-or-none" matching (Proposal). In this table, we observe that the initial matching produces images with better Hausdorff distance than the image provided by the DNN or the proposal but with smaller Dice index in some cases (e.g. U-Net with the B training configuration in Table \ref{tab:fasseg_ablation}). By contrast, the proposal images have, to some extent, a better Dice index than the ones provided by the DNN or the initial matching. These results illustrate the fact that our initial matching recovers for each class a proper region and that our refinement operation aggregates new regions to those regions,  which can increase the Dice index but also the Hausdorff distance.

\begin{table}[!ht]
\caption{
Results on the FASSEG dataset, for different DNNs and training configurations. Results are given as Dice score (DSC) and Hausdorff distance (HD) with standard deviation in parentheses for the DNN output (DNN), the initial “one-to-one-or-none”  matching (Inter) and the “many-to-one-or-none”  matching (Proposal).}
\label{tab:fasseg_ablation}
\resizebox{\linewidth}{!}{%
\begin{tabular}{|c|c|c|c|c|c|c|}

\hline
Training & \multicolumn{6}{|c|}{A training configuration}\\

\hline
Method & \multicolumn{2}{|c}{DNN} & \multicolumn{2}{|c}{Inter} & \multicolumn{2}{|c|}{Proposal}\\

\hline
Model & DSC$\uparrow$  & HD$\downarrow$ & DSC$\uparrow$  & HD$\downarrow$ & DSC$\uparrow$  & HD$\downarrow$\\

\hline

U-Net (2015) & 0.7 (0.05)& 37.4 (19.32)& 0.69 (0.06)& \textbf{27.36} (12.92)& 0.7 (0.05)& 35.79 (21.33)\\

\hline

U-Net + CRF (2015) & 0.71 (0.05)& 35.4 (17.18)& 0.71 (0.05)& \textbf{24.54} (10.63)& 0.71 (0.05)& 34.07 (18.21)\\

\hline

PSPNet (2017) & 0.83 (0.05)& 24.18 (16.58)& 0.81 (0.08)& 22.45 (15.83)& 0.83 (0.06)& \textbf{20.44} (14.82)\\

\hline

EfficientNet (2019) & 0.84 (0.05)& 22.11 (17.0)& 0.83 (0.06)& \textbf{18.28} (13.18)& 0.84 (0.05)& 20.48 (17.04)\\

\hline

\end{tabular}
}

\bigskip

\resizebox{\linewidth}{!}{%
\begin{tabular}{|c|c|c|c|c|c|c|}

\hline
Training & \multicolumn{6}{|c|}{B training configuration}\\

\hline
Method & \multicolumn{2}{|c}{DNN} & \multicolumn{2}{|c}{Inter} & \multicolumn{2}{|c|}{Proposal}\\

\hline
Model & DSC$\uparrow$  & HD$\downarrow$ & DSC$\uparrow$  & HD$\downarrow$ & DSC$\uparrow$  & HD$\downarrow$\\

\hline

U-Net (2015) &0.72 (0.06)& 43.02 (26.64)& 0.72 (0.06)& \textbf{26.05} (12.92)& \textbf{0.73} (0.05)& 32.59 (17.73)\\

\hline

U-Net + CRF (2015) & 0.7 (0.05)& 39.87 (20.71)& 0.7 (0.06)& \textbf{26.88} (12.49)& \textbf{0.71} (0.05)& 34.48 (20.31)\\

\hline

PSPNet (2017) & 0.82 (0.05)& 27.8 (23.0)& 0.81 (0.07)& 21.93 (15.04)& 0.82 (0.05)& \textbf{20.96} (15.63)\\

\hline

EfficientNet (2019) & 0.83 (0.05)& 26.39 (18.51)& 0.82 (0.06)& \textbf{20.53} (14.82)& 0.83 (0.05)& 22.64 (18.82)\\

 \hline

\end{tabular}
}

\bigskip

\resizebox{\linewidth}{!}{%
\begin{tabular}{|c|c|c|c|c|c|c|}

\hline
Training & \multicolumn{6}{|c|}{C training configuration}\\

\hline
Method & \multicolumn{2}{|c}{DNN} & \multicolumn{2}{|c}{Inter} & \multicolumn{2}{|c|}{Proposal}\\

\hline
Model & DSC$\uparrow$  & HD$\downarrow$ & DSC$\uparrow$  & HD$\downarrow$ & DSC$\uparrow$  & HD$\downarrow$\\

\hline

U-Net (2015) & 0.71 (0.07)& 54.13 (33.5)& 0.72 (0.07)& \textbf{28.66} (15.75)& 0.72 (0.07)& 37.55 (22.27)\\

\hline

U-Net + CRF (2015) & 0.69 (0.06)& 60.01 (30.55)& 0.67 (0.1)& \textbf{37.34} (21.3)& 0.69 (0.08)& 43.99 (26.14)\\

\hline

PSPNet (2017) & 0.76 (0.09)& 79.93 (47.03)& 0.72 (0.14)& 35.35 (25.68)& 0.76 (0.11)& \textbf{33.07} (25.32)\\

\hline

EfficientNet (2019) & 0.81 (0.05)& 49.8 (33.88)& 0.8 (0.07)& \textbf{22.17} (15.35)& 0.81 (0.06)& 29.13 (23.27)\\

\hline

\end{tabular}
}
\end{table}

Table \ref{tab:FASSEG_results} reports detailed results for each class, for the B training configuration with the PSPNet (for the sake of brevity, results are not detailed for all DNNs and training configurations, and details for other training configurations and DNN backbones are provided in \ref{sec:appendixFASSEG}). First, it appears, as in Table~\ref{tab:FASSEG_results_globale}, that the improvement is significant in terms of Hausdorff Distance, as previously underlined. Our method significantly improves the segmentation, except for hair (Hr) where the Hausdorff distance is slightly worse than with PSPNet only. This results from the fact that the relative position of hair, with respect to other structures, varies from one person to another one (e.g. short hair or long hair), compared to other structures, this having an impact on the graph model ($G_m$) and therefore on the efficiency of the graph matching.

\begin{table}[!ht]
\centering
\caption{Example of detailed results on the FASSEG dataset, with the use of PSPNet and the B training configuration (see Table \ref{tab:FASSEG_results_globale}). Results, averaged over 3 random selections of the training dataset, are given in terms of Dice score (DSC) and Hausdorff distance (HD), for each class: Hr (hair), Fc (face), L-br (left eyebrow), R-br (right eyebrow), L-eye (left eye), R-eye (right eye), nose and mouth.
}
\label{tab:FASSEG_results}
\begin{tabular}{|c|c|c|c|c|}

\hline
Method & \multicolumn{2}{|c}{PSPNet} & \multicolumn{2}{|c|}{Proposal} \\

\hline
Class & DSC$\uparrow$  & HD $\downarrow$ & DSC $\uparrow$ & HD $\downarrow$ \\

\hline

Hr & \textbf{0.93} & {\bf 72.48} & 0.92 & 77.62 \\
Fc & 0.95 & 30.73 & 0.95 & \textbf{24.08} \\
L-br & 0.72 & 16.08 & \textbf{0.73} & \textbf{13.60} \\
R-br & 0.71 & 20.37 & 0.71 & \textbf{16.09} \\
L-eye & 0.85 & 11.03 & 0.85 & \textbf{8.14} \\
R-eye & 0.83 & 22.30 & 0.83 & \textbf{12.54} \\
Nose & 0.76 & 26.59 & 0.76 & \textbf{9.69} \\
Mouth & 0.85 & 22.93 & 0.85 & \textbf{9.34} \\
\hline
Mean & 0.82 & 27.82 & \textbf{0.83} & \textbf{21.39} \\
\hline
\end{tabular}
\end{table}

\begin{figure*}[!h]
    \centering
    \includegraphics[width = 0.9\textwidth]{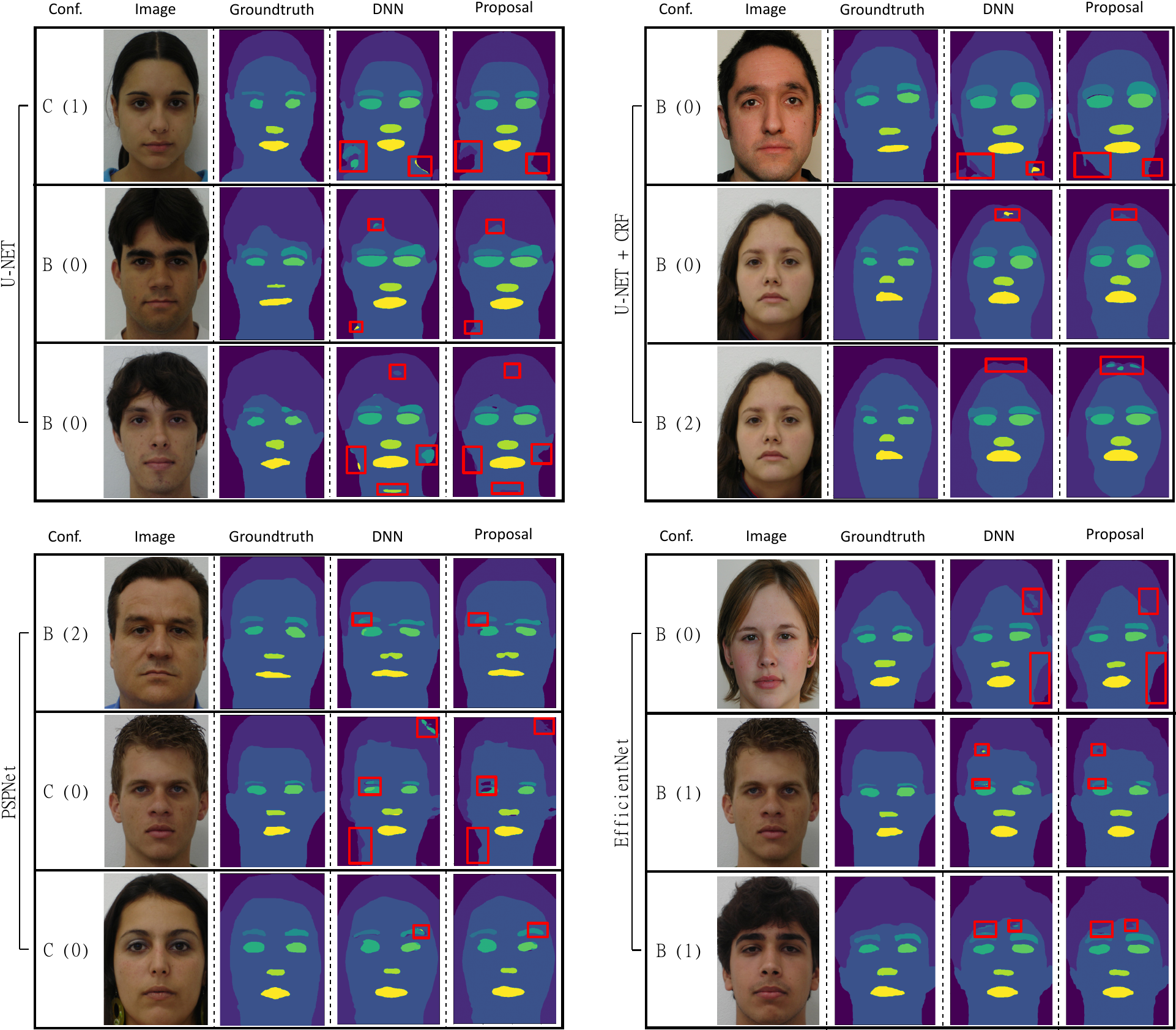}
    \caption{Example of segmentation results for various DNN and training configuration (the index of the random selection of the training dataset, over the 3 considered, is reported between parenthesis).}
    \label{fig:comparasion_face}
\end{figure*}

Figure \ref{fig:comparasion_face} gives some examples of semantic segmentations for various DNNs and training configurations (one favors B and C configurations for which the improvement is more significant). These examples show the benefits of integrating high-level structural information from the annotated dataset to ensure the spatial coherence of the final segmentation, by appropriately relabelling some segmented regions initially incorrectly labelled by the DNN (see surrounded regions). Our method is robust with respect to the selection of the training dataset. This is illustrated by Figure~\ref{fig:comparasion_face}-U-Net+CRF with two different segmentations of the same face (bottom B(0) and B(2) images), based on two different random selections of training images: in both cases, our method appropriately corrects structural errors. In some cases, it may not appropriately correct the DNN-based segmentation, as illustrated with PSPNet in the middle image C(0). In this example, the DNN makes two mistakes regarding the left eyebrow and eye, corresponding to two regions that are finally discarded by our approach instead of being appropriately relabelled. This is directly due to the hyperparameters of our approach: the threshold $T$ in Equation \ref{eq:T} (for discarding regions) and the coefficient balancing the contribution of node and edge attributes ($\lambda $ in Equation \ref{eq:K}), as well as the one balancing relationships ($\lambda_e$ in Equation \ref{D_edt}).

Note that, in some cases, the neural network may have difficulties to identify some classes, due to a poor representativity of the training dataset (C training configuration). This leads to a preliminary semantic segmentation where some classes are missing, and this is not handled by our approach: the first step (``one-to-one'' matching) assumes that the output of DNN contains at least one region per class. In our experiments, when the DNN failed finding some regions, related images have been discarded.

\subsection{IBSR}
\label{sec_ibsr}
\subsubsection{Data}
The IBSR\footnote{The IBSR annotated public dataset can be downloaded at the following address: \url{https://www.nitrc.org/projects/ibsr}.} public dataset provides 18 3D MRI of the brain, together with the manual segmentation of 32 regions. In our experiments, similarly to \cite{KUSHIBAR2018177},  only 14 classes (i.e. 14 regions) of the annotated dataset are considered: thalamus (left and right - Tha.L and Tha.R), caudate (left and right - Cau.L and Cau.R), putamen (left and right - Put.L and Put.R), pallidum (left and right - Pal.L and Pal.R), hippocampus (left and right - Hip.R and Hip.L), amygdala  (left and right - Amy.L and Amy.R) and accumbens (left and right - Acc.L and Acc.R).

\subsubsection{Structural information}
The model graph $G_m$ is a complete graph with $14$ vertices corresponding to the $14$ classes (see Figure \ref{fig:Gms}-IBSR).

Vertex attributes include
both the class membership probability vector, as for FASSEG (Section~\ref{sec:structFASSEG}), and a region property corresponding to the largest distance between two points belonging to the same region. This property is similar to the $d_{max}^{(i,j)}$ distance between regions (Equation~\ref{edt_max}), 
but computed here over points in the same region:
\begin{equation}
d_{max}^i= \frac{\max_{ (p,q) \in R_i^2 }(|p-q|)}{C_s}    
\end{equation}
As previously, $C_s$ denotes the maximum distance value observed in an image, ensuring that values range within $[0,1]$.
This region attribute is considered to favor compact regions. To compute $K_v$, we consider the dissimilarity function ${{}_vD_2}_{i}^{k}$ ($i\in V_r$ and $k\in V_m$):
\begin{equation}
\label{A_pos}
{{}_vD_2}_{i}^{k}=\lambda_v \frac{1}{N}\sum_{1\leq n\leq N}(S_{m,k}[n]-S_{r,i}[n])^2 + (1-\lambda_v) (\left| d_{max}^k-d_{max}^i\right|) 
\end{equation}
The term $\lambda_v$ is a parameter balancing
the influence of the class membership probability and the size of the region.

The edge attribute, corresponding to a spatial relationship, is the relative directional position of the centroid of two regions, as in \cite{Bloch2012}. For two regions $R_i$ and $R_j$, the relative position, related to the function $\alpha_e$, is defined by a vector 
\begin{equation}
\forall e=(i,j)\in E_{.},\ \alpha_e(e) = \Vec{v_{ij}}  = \frac{\overline{R}_j-\overline{R}_i}{C_s}    
\end{equation}
where $E_.$ is either $E_r$ (graph $G_r$) or $E_m$ (graph $G_m$). The term $\overline{R}$ denotes the coordinates of the center of mass of region $R$. 
Based on this relationship, the matrix $K_e$ is computed using the following dissimilarity function:
\begin{equation}
\label{D_pos}
{{}_eD_2}_{(i,j)}^{(k,l)}=\lambda_e \frac{|\cos \theta -1|}{2} + (1-\lambda_e) \frac{\left| \|\Vec{v_{ij}}\|_2 - \|\Vec{v_{kl}}\|_2 \right|}{C_s}  
\end{equation}
where $\theta$ is the angle between them $\vec{v_{ij}}$ and $\vec{v_{kl}}$ vectors, computed using a scalar product:
\begin{equation}
    \cos (\theta) = \frac{\Vec{v_{ij}}.\Vec{v_{kl}}}{\|\Vec{v_{ij}}\|_2.\|\Vec{v_{kl}}\|_2}
    \label{D_scalar}
\end{equation}
The term $\lambda_v \in [0, 1]$ (respectively $\lambda _e$) is a parameter balancing the influence of the difference in terms of class membership probability and region size (respectively the influence of the difference in terms of distance and orientation).

As mentioned in Section \ref{subsec:graphhypo}, model graphs are computed by considering annotated images of the training dataset (mean distances and relative orientations between regions).

\subsubsection{Hyperparameters}

For the experiments on the IBSR dataset, the values of $\lambda$ (Equation \ref{eq:K}), $\lambda_v$ (Equation \ref{A_pos}) and $\lambda_e$ (Equation~\ref{D_pos}) are all set to 0.5.
The value of $T$ in Equation \ref{eq:T} is set to $1$. 
Those parameters are chosen empirically and no optimization has been applied. 
The same parameter values have been used for all the DNN backbones without any specific tuning.

\subsubsection{DNN backbones}
For this dataset, we consider several DNNs: a 3D U-Net neural network \cite{3DU-Net2016}, a 3D U-Net combined with CRF \cite{CRF2015} and the recent UNETR based on transformers \cite{2022_Unetr}. Note that, except for the U-Net, DNNs are different from those considered for FASSEG because one faces 3D images (while for FASSEG, available implementations of considered DNNs are designed for 2D color images). A specificity of such 3D medical images is the size and the fact that classes are highly unbalanced, requiring specific strategies such as the use of patches.

The 3D U-Net neural network is based on the  generic 2D/3D implementation also considered for FASSEG \cite{codeUnet_2019_3522306}, including a comparison with a CRF-based post-processing \cite{CRF2015}, as for FASSEG (with the same Gaussian filter size). 60 epochs are used for training the network and an early stopping strategy is applied to prevent over-fitting. The training process is terminated if their is no improvement on the loss (using cross entropy loss function) for 8 consecutive epochs. We consider a 3D patch-based approach~\cite{patch_training}, this being motivated by the fact that classes are highly unbalanced (i.e. small size of target regions with respect to other brain tissues and background). Patches are volumes of size $32^3$ voxels, that are extracted around the centroid of each label (random selection) using the \emph{Torchio} library \cite{perez-garcia_torchio_2020}. 64 patches are selected for each MRI image, with a frequency that is proportional to the inverse prior probability of the corresponding class. 

We also consider the recent UNETR \cite{2022_Unetr}, based on transformers, and designed for medical image segmentation. Such a network, based on the attention mechanism, allows modeling long-range (spatial) dependencies and capturing global context, this not being the case with standard DNNs (context limited to the localized receptive field) \cite{2022_Unetr}.

In our experiments, the hyperparameters have been set as in~\cite{2022_Unetr}, with patches of size $16^3$.

The number of considered training, validation and test images is reported in Table \ref{tab:datasetsize}, for the various configurations.

\begin{table}[!ht]
\begin{center}

\small
\caption{
Number of trainable parameters for each architectures used for the IBSR dataset.}
\label{tab:IBSR_parameters}

\begin{tabular}{|c|c|}

\hline
Architectures & Nb. trainable parameters \\
\hline
U-NET & 15 711 887\\
\hline
U-NET + CRF & 15 711 891\\
\hline
UNETR & 94 197 951\\
\hline

\end{tabular}
\end{center}
\end{table}

\subsubsection{Results}
\label{sec:IBSR_results}

Table \ref{tab:IBSR_results_globale} provides segmentation results for different DNN backbones, as well as for different sizes of the training configurations. Note that, for the C training configuration using UNETR, the measured efficiency is only based on one trained network (i.e. one random selection of training images) instead of 3, as considered for other experiments (see evaluation protocol described in Section~\ref{subsec:proto}). Any other tested randomly selected training images did not enable to correctly train the UNETR in order to find, on test images, at least one region candidate for each of the 14 classes, leading to very bad segmentation results. This poor results are therefore discarded in Table \ref{tab:IBSR_results_globale}. This illustrates that such a transformer-based DNN requires a larger training dataset than other DNNs (i.e. UNet in our case). Concerning results reported in this table, the improvement appears significant in terms of Hausdorff distance as for FASSEG. This underlines the relevance of our approach to correct structural errors (i.e. parts of regions being located far away from the target, leading to a structural incoherence). It is the case, in particular, for the UNETR, based on transformer and therefore the attention mechanism trying to take into account, to a certain extent, the spatial information to guide the segmentation (UNETR performs significantly better that the U-Net in terms of HD, except for C training configuration, as previously underlined). In this case, our approach significantly reduces the HD from $11.01$ to $5.4$ (reduction of $51$\%) for the A training configuration, from $13.76$ to $5.52$ (reduction of $59.8$ \%) for the B training configuration and from $25.8$ to $7.13$ (reduction of $72.3$) for the C training configuration. As for FASSEG, this also illustrates the ability of our method to significantly compensate the lack of representativity when the training dataset is small (trend illustrated in Figure~\ref{fig:hd_config_curves}-IBSR): relationships observed at high level in the annotated dataset, and exploited by our approach, allow us to correct structural mistakes achieved by a DNN, even with the use of the attention mechanism. 

\begin{table*}[!ht]
\caption{
Results on the IBSR dataset, for different DNNs and training configurations. Results are given in terms of Dice score (DSC) and Hausdorff distance (HD) with average results computed using results from 18 test images (the 6 test images with each DNN trained 3 times) and standard deviation in parentheses, together with minimum and maximum values (between brackets)}
\label{tab:IBSR_results_globale}
\resizebox{\textwidth}{!}{%
\begin{tabular}{|c|c|c|c|c|c|c|c|c|c|c|c|c|}

\hline
Tr. dataset & \multicolumn{4}{|c}{A} & \multicolumn{4}{|c}{B} & \multicolumn{4}{|c|}{C}\\

\hline
Method & \multicolumn{2}{|c}{DNN} & \multicolumn{2}{|c}{Proposal} & \multicolumn{2}{|c}{DNN} & \multicolumn{2}{|c}{Proposal} & \multicolumn{2}{|c}{DNN} & \multicolumn{2}{|c|}{Proposal}\\

\hline
Model & DSC$\uparrow$  & HD$\downarrow$  & DSC$\uparrow$  & HD$\downarrow$ & DSC$\uparrow$  & HD$\downarrow$ & DSC$\uparrow$  & HD$\downarrow$ & DSC$\uparrow$  & HD$\downarrow$ & DSC$\uparrow$  & HD$\downarrow$ \\

\hline

\multirow{2}{*}{U-Net}& 0.81 (0.02)& 25.82 (6.93)& \textbf{0.82} (0.02)& \textbf{4.61} (0.49)& 0.79 (0.03)& 22.99 (7.81)&\textbf{ 0.81} (0.02)& \textbf{4.74} (0.51)& 0.76 (0.03)& 26.94 (8.37)& \textbf{0.78} (0.03)&\textbf{ 5.19} (0.87)\\
& [0.76; 0.84] & [14.04; 39.43]& [0.78; 0.85] & [3.46; 5.43]& [0.74; 0.84] & [8.6; 36.91]& [0.75; 0.84] & [3.72; 5.59]& [0.69; 0.82] & [14.03; 45.38]& [0.72; 0.83] & [4.09; 7.24]\\

\hline

\multirow{2}{*}{U-Net + CRF}& 0.81 (0.02)& 25.82 (6.93)& \textbf{0.82} (0.02)& \textbf{4.61} (0.49)& 0.79 (0.02)& 23.49 (7.76)& \textbf{0.81} (0.02)& \textbf{4.73} (0.65)& 0.77 (0.04)& 28.5 (8.43)& \textbf{0.8} (0.03)& \textbf{5.22} (1.35)\\
& [0.76; 0.84] & [14.04; 39.43]& [0.78; 0.85] & [3.46; 5.43]& [0.74; 0.83] & [9.65; 39.29]& [0.77; 0.84] & [3.7; 6.06]& [0.7; 0.83] & [13.39; 46.73]& [0.74; 0.84] & [3.64; 8.63]\\

\hline

\multirow{2}{*}{UNETR}& 0.75 (0.02)& 11.01 (3.75)& 0.75 (0.02)& \textbf{5.4} (0.74)& 0.73 (0.03)& 13.76 (3.9)& 0.73 (0.03)& \textbf{5.52} (0.75)& 0.69 (0.03)& 24.09 (1.02)& 0.69 (0.03)& \textbf{6.42} (1.15)\\
& [0.71; 0.79] & [4.97; 17.67]& [0.71; 0.79] & [4.4; 6.83]& [0.67; 0.78] & [6.8; 22.71]& [0.67; 0.78] & [4.49; 7.02]& [0.66; 0.72] & [22.33; 24.79]& [0.67; 0.73] & [5.25; 8.25]\\
\hline

\hline

\end{tabular}
}
\end{table*}

An interesting point concerns the number of trainable parameters (see Table \ref{tab:IBSR_parameters}): without our proposal, UNETR performs better than the U-Net in terms of HD, but at the cost of a significantly larger number of parameters. Thanks to our approach, the HD is significantly reduced in the case of the U-Net which becomes more efficient than the UNETR (both in terms of Dice score and HD), with less parameters. This illustrates the relevance of considering high-level relationships available on the annotated dataset, not only in terms of segmentation efficiency, but also in terms of number of trainable parameters, this being a challenging issue, as recently underlined~\cite{2022_convnet}. Note that 210 parameters are computed from the annotated dataset to build the graph model $G_m$ (relationships between the 14 classes, together with vertex attributes related to the region size). Table~\ref{tab:nb_vertices_ibsr} reports the number of vertices in $G_r$, observed in our experiments, to be matched with those of the model graph $G_m$ ($|V_m|=14$ as illustrated in Figure~\ref{fig:Gms}). This number increases as the training dataset size decreases (this being particularly significant with the UNETR DNN).

\begin{table}[!ht]
\caption{
Average number of regions ($\overline{\#R}$) in the segmentation provided by the DNNs considered for the IBSR dataset (the ideal number of regions is 14) and the running time (Exec time in second) of our method  of our graph matching method to process those segmentation maps. The average number of regions and the average execution time ($\overline{\text{Exe}}$) considering each training configuration are reported with standard deviation (in parenthesis).}
\label{tab:nb_vertices_ibsr}
\resizebox{\linewidth}{!}{  
\begin{tabular}{|c|c|c|c|c|}
\hline
Model & Metrics &  Configuration A & Configuration B & Configuration C\\

\hline

\multirow{2}{*}{U-Net (2015)} & $\overline{\#\text{R}}$ & 33.83 (7.24) & 37.33 (19.35) & \pmb{36.72} (10.23) \\
 & $\overline{\text{Exe}}$ & 218.02 (77.47) & 260.80 (214.18) &  \pmb{254.26} (119.12) \\

\hline

\multirow{2}{*}{U-Net + CRF (2015)} & $\overline{\#\text{R}}$ & 33.83 (7.24) & \pmb{33.44} (9.32) & 42.72 (13.67)  \\
 & $\overline{\text{Exe}}$ & 218.43 (77.93) & \pmb{214.99} (100.34) & 330.66 (175.47) \\

\hline

\multirow{2}{*}{UNETR (2022)} & $\overline{\#\text{R}}$ & \pmb{29.06} (9.74) & 52.33 (40.30) & 77.0 (2.23) \\
 & $\overline{\text{Exe}}$ & \pmb{167.71} (103.49) & 2169.14 (3969.37) & 4870.28 (1488.69) \\

\hline
\end{tabular}
}
\end{table}

Note that, compared to a recent work on this dataset~\cite{KUSHIBAR2018177} but involving an atlas (our approach has the strength of being atlas free), our approach leads to slightly worse results in terms of Dice score ($0.84$ in~\cite{KUSHIBAR2018177}) but to a similar efficiency in terms of Hausdorff distance ($4.49$ in~\cite{KUSHIBAR2018177}). 

Table \ref{tab:IBSR_ablation} reports the results we obtained for an ablation study where we compare the results of the segmentation after the initial "one-to-one-or-none" matching (Inter) and after the refinement operation leading to a "many-to-one-or-none" matching (Proposal). In this table, we observe that the initial matching produces images with better Hausdorff distance than the image provided by the DNN or the proposal. Moreover, the initial matching has better results in terms of Dice index than the image provided by the DNN and the same results as the proposal images. However, the proposal results are better than the output of the DNN and the difference between the proposal and the initial matching are small and might not be significant in regards to the standard deviation. These results illustrate the fact that our initial matching recovers for each class a proper region and that our refinement operation  aggregates new small regions to these regions,  which increases the Hausdorff distance but not the Dice index. This can be, to some extent, explained by the fact the DNNs provides one region of interest for each class and artifacts around it thus the merging operation does not provide interesting benefit in this context.

\begin{table}[!ht]
\caption{
Results on the IBSR dataset, for different DNNs and training configurations. Results are given as Dice score (DSC) and Hausdorff distance (HD) with standard deviation in parentheses for the DNN output (DNN), the initial “one-to-one-or-none”  matching (Inter) and the “many-to-one-or-none”  matching (Proposal).}
\label{tab:IBSR_ablation}
\resizebox{\linewidth}{!}{%
\begin{tabular}{|c|c|c|c|c|c|c|}

\hline
Training & \multicolumn{6}{|c|}{A training configuration}\\

\hline
Method & \multicolumn{2}{|c}{DNN} & \multicolumn{2}{|c}{Inter} & \multicolumn{2}{|c|}{Proposal}\\

\hline
Model & DSC$\uparrow$  & HD$\downarrow$ & DSC$\uparrow$  & HD$\downarrow$ & DSC$\uparrow$  & HD$\downarrow$\\

\hline
U-Net& 0.81 (0.02)& 25.82 (6.93)& 0.82 (0.02)& \textbf{4.4} (0.48)& 0.82 (0.02)& 4.61 (0.49)\\
\hline
U-Net + CRF & 0.81 (0.02)& 25.82 (6.93)& 0.82 (0.02)& \textbf{4.4} (0.48)& 0.82 (0.02)& 4.61 (0.49)\\
\hline
UNETR& 0.75 (0.02)& 11.01 (3.75)& 0.75 (0.02)& \textbf{5.31} (0.74)& 0.75 (0.02)& 5.4 (0.74)\\
\hline

\end{tabular}
}

\bigskip

\resizebox{\linewidth}{!}{%
\begin{tabular}{|c|c|c|c|c|c|c|}

\hline
Training & \multicolumn{6}{|c|}{B training configuration}\\

\hline
Method & \multicolumn{2}{|c}{DNN} & \multicolumn{2}{|c}{Inter} & \multicolumn{2}{|c|}{Proposal}\\

\hline
Model & DSC$\uparrow$  & HD$\downarrow$ & DSC$\uparrow$  & HD$\downarrow$ & DSC$\uparrow$  & HD$\downarrow$\\

\hline

U-Net& 0.79 (0.03)& 22.99 (7.81)& 0.81 (0.02)& \textbf{4.62} (0.54)& 0.81 (0.02)& 4.74 (0.51)\\
\hline
U-Net + CRF& 0.79 (0.02)& 23.49 (7.76)& 0.81 (0.02)& \textbf{4.57} (0.55)& 0.81 (0.02)& 4.73 (0.65)\\
\hline
UNETR& 0.73 (0.03)& 13.76 (3.9)& 0.73 (0.03)& \textbf{5.38} (0.74)& 0.73 (0.03)& 5.52 (0.75)\\
\hline

\end{tabular}
}

\bigskip

\resizebox{\linewidth}{!}{%
\begin{tabular}{|c|c|c|c|c|c|c|}

\hline
Training & \multicolumn{6}{|c|}{C training configuration}\\

\hline
Method & \multicolumn{2}{|c}{DNN} & \multicolumn{2}{|c}{Inter} & \multicolumn{2}{|c|}{Proposal}\\

\hline
Model & DSC$\uparrow$  & HD$\downarrow$ & DSC$\uparrow$  & HD$\downarrow$ & DSC$\uparrow$  & HD$\downarrow$\\

\hline

U-Net& 0.76 (0.03)& 26.94 (8.37)& 0.78 (0.03)& \textbf{5.07} (0.77)& 0.78 (0.03)& 5.19 (0.87)\\
\hline
U-Net + CRF& 0.77 (0.04)& 28.5 (8.43)& 0.8 (0.03)& \textbf{4.79} (0.8)& 0.8 (0.03)& 5.22 (1.35)\\
\hline
UNETR& 0.69 (0.03)& 24.09 (1.02)& 0.69 (0.03)& \textbf{5.72} (1.12)& 0.69 (0.03)& 6.42 (1.15)\\
\hline

\end{tabular}
}
\end{table}

Table \ref{tab:IBSR_results} reports detailed results for each class, for the B training configuration with the U-Net together with CRF (for sake of brevity, results are not detailed for all DNNs and training configurations, and details for other training configurations and DNN backbones are provided in \ref{sec:appendixIBSR}). First, it appears, as in Table \ref{tab:IBSR_results_globale}, that the improvement appears significant in terms of Hausdorff distance, as previously underlined. Our method significantly improves the segmentation, except for left and right accumbens (similar HD). In some cases, the Dice score is also improved: left thalamus (6$\%$), right thalamus (7$\%$), left hippocampus (8$\%$) and right hippocamus (4$\%$). This illustrates that our method can also significantly improve the segmentation in terms of Dice score for some classes, even if it appears negligible in average (see Table~\ref{tab:IBSR_results_globale}).

\begin{table}[!ht]
%\large
\centering
\caption{Example of detailed results on the IBSR dataset, with the use of U-Net together with CRF, and the B training configuration (see Table \ref{tab:IBSR_results_globale}). Results, averaged over 3 random selections of the training dataset, are given in terms of Dice score (DSC) and Hausdorff distance (HD), for each class: Tha.L(left thalamus), Tha.R(right thalamus), Cau.L(left caudate), Cau.R(right caudate), Put.L(left putamen), Put.R(right putamen), Pal.L(left pallidum), Pal.R(right pallidum), Hip.L(left hippocampus), Hip.R(right hippocamus), Amy.L(left amygdala), Amy.R(right amygdala), Acc.L(left accumbens), Acc.R(right accumbens).}
\label{tab:IBSR_results}

\begin{tabular}{|c|c|c|c|c|}

\hline
Method & \multicolumn{2}{|c}{DNN} & \multicolumn{2}{|c|}{Proposal} \\

\hline
Class & DSC$\uparrow$  & HD$\downarrow$  & DSC$\uparrow$  & HD$\downarrow$ \\

\hline

Tha.L & 0.84 & 39.76 & \textbf{0.9} & \textbf{5.5} \\
Tha.R & 0.83 & 40.86 & \textbf{0.9} & \textbf{4.02} \\
Cau.L & 0.84 & 18.42 & 0.84 & \textbf{5.87} \\
Cau.R & 0.85 & 16.91 & 0.85 & \textbf{4.49} \\
Put.L & 0.88 & 13.74 & 0.88 & \textbf{4.59} \\
Put.R & 0.88 & 12.07 & 0.88 & \textbf{4.77} \\
Pal.L & 0.82 & 7.24 & 0.82 & \textbf{3.72} \\
Pal.R & 0.77 & 6.93 & 0.77 & \textbf{4.28} \\
Hip.L & 0.74 & 64.75 & \textbf{0.82} & \textbf{5.48} \\
Hip.R & 0.79 & 64.27 & \textbf{0.83} & \textbf{5.72} \\
Amy.L & 0.73 & 14.69 & 0.73 & \textbf{4.94} \\
Amy.R & 0.76 & 21.31 & 0.76 & \textbf{4.93} \\
Acc.L & 0.67 & 3.72 & 0.67 & \textbf{3.72} \\
Acc.R & 0.63 & 4.18 & 0.63 & \textbf{4.18} \\
\hline
Mean & 0.79 & 23.49 & \textbf{0.81} & \textbf{4.73} \\
\hline
\end{tabular}

\end{table}

\begin{figure*}[!ht]
    \centering
    \includegraphics[height = 0.9\textheight]{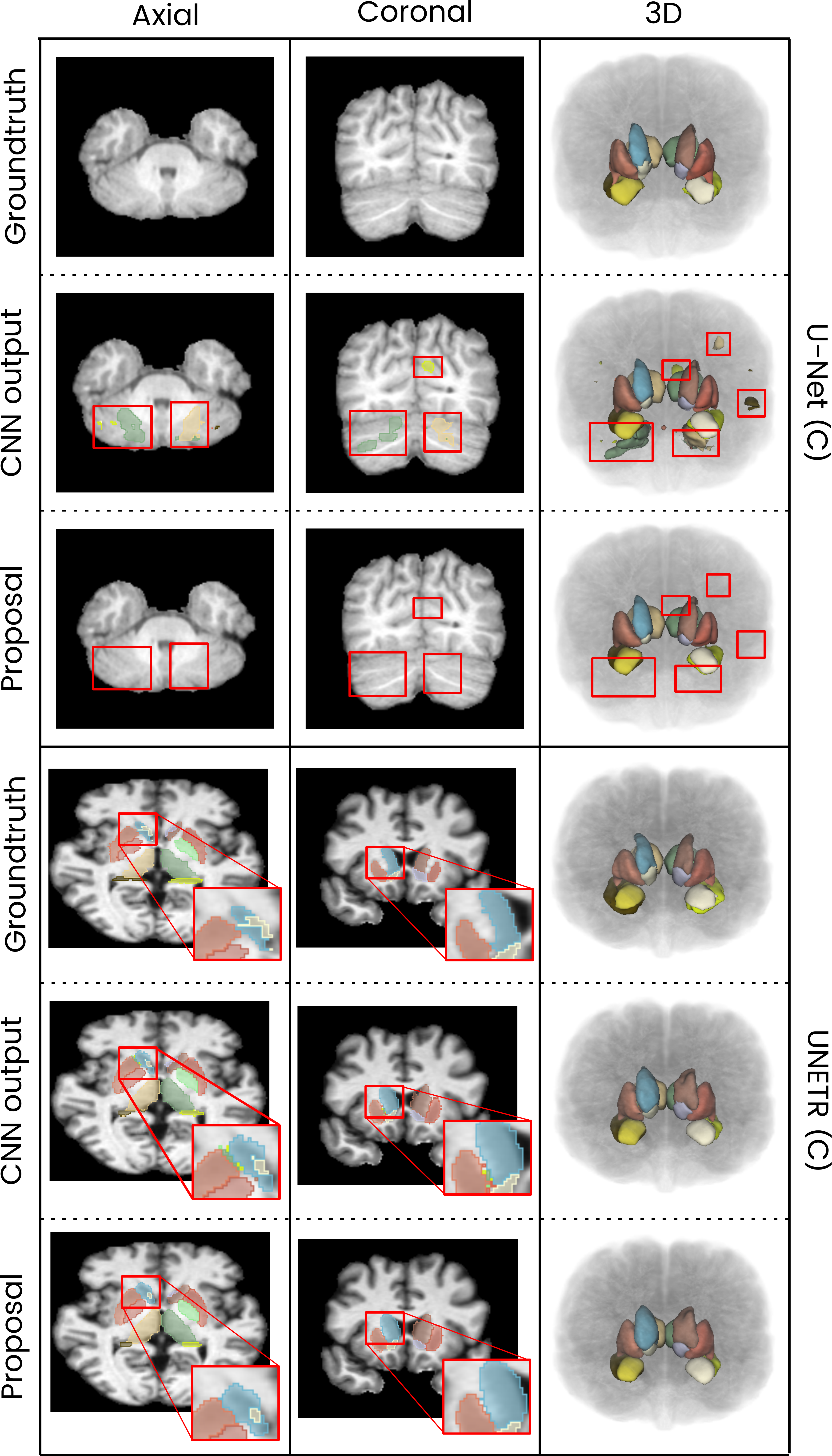}
    \caption{Two examples of results on the IBSR dataset, in the case of the C training configuration and for both U-Net and UNETR DNN backbones. Surrounded boxes indicate some regions that have been appropriately relabelled.}
    \label{fig:comparasion_brain}
\end{figure*}

Figure \ref{fig:comparasion_brain} provides two examples of 3D images for two DNN backbones (U-Net and UNETR) in the C training configuration (configuration for which the improvement is visually more significant). The U-Net produces many errors corresponding to segmented regions standing far away from the target one (leading to a high HD as illustrated in Table~\ref{tab:IBSR_results_globale}). 
Our approach enables to remove such artefacts, corresponding to vertices that are relabelled as ``none'', the underlying regions being merged with the background. This is managed during the refinement according to the removal threshold ($T$ in Equation \ref{eq:T} and in Algorithm \ref{algo_qap}). 
Concerning the other image example, UNETR does not produce such errors, ensuring a better spatial coherence resulting from the attention mechanism. 
The observed errors correspond to some regions, close to the target ones, that are initially incorrectly classified (see surrounded boxes in Figure~\ref{fig:comparasion_brain}-UNETR). In such a case, our method correctly relabels such vertices and the corresponding  regions that merged, so that the structural coherence is preserved.

\section{Discussion and perspectives}
\label{sec:discussion}

Both our experiments demonstrate that the use of structural relationships  improves the semantic segmentation provided by neural networks. 
The nature of the structural information encoded in our graphs  is a hyperparameter in our approach. 
These graphs need to be appropriately designed for the targeted application.

We have also tested  the structural information presented for IBSR on FASSEG but it did  not lead to good results on FASSEG. Similarly, the structural information considered for IBSR, based on the relative directional position of region barycenters, is not appropriate for FASSEG. For instance, the center of mass of hair may strongly vary from one person to another one (strongly varying concave region), leading to strong variations in terms of relative positions, and involving graph matching errors. Hence well chosen information encoded as vertices and edges, and their  associated dissimilarities allow improving the baseline DNNs results.

In our experiments, structural models  (cf. graph models in Figure \ref{fig:Gms}) have been
designed from the annotated dataset as a  \textit{commonsense graph} \cite{Giuliari_2022_CVPR} in the context of our targeted applications.
These have been shown to be  appropriate for our applications where the overall structure is relatively stable from one image to another such as  the structure of the face (e.g. FASSEG) and of the body (e.g. IBSR for medical imaging).
We believe that other applications of our approach could prove likewise successful for segmenting well structured patterns such as those found in  handwritten documents (e.g. handwritten forms) or for segmenting geographic information (e.g. remote sensing imagery \cite{remotesensing}). 
Future work will aim at considering other datasets (e.g. CelebAMask-HQ \cite{CelebAMask-HQ}, Cityscapes \cite{2017_PSPNet,ContextPrior2020, JIN202129}, RUGD \cite{ContextPrior2020,JIN202129}, ADE20K \cite{2017_PSPNet, 2021_segFormer}, PASCAL-Context \cite{ContextPrior2020}, Pascal VOC 2012 \cite{2017_PSPNet}, BTCV and MSD \cite{2022_Unetr}). Note that it will involve potentially more complex graph models to manage the strong structural variations observed in such datasets. This is out of the scope of the paper showing the relevance of our method, with various recent DNN backbones, on applications with relatively stable structural information. 

A strength of our approach is to directly integrate relationships observed in annotated datasets (note that it could also be provided by qualitative domain knowledge such as in \cite{fasquel2017}). Such knowledge integration can be done without
any particular tuning of the initial DNN, compared to approaches attempting to implicitly integrate such (spatial) information within the neural network, at the cost of the increase of its complexity, in particular in terms of number of trainable parameters (additionally involving the requirement of a large training dataset) \cite{2022_convnet}. In such a case (i.e. integration of spatial information directly in the DNN), the neural network architectures must also be adapted according to the ranges of spatial relationships to be considered for a given application. This involves, for instance, the appropriate selection of the number of layers or transformer blocks for the UNETR we considered in the experiments \cite{2022_Unetr}, the size of the convolution kernel in the aggregation module of the CPNet \cite{ContextPrior2020}, related to the notion of receptive field capturing spatial relationships. The same comment can be done regarding the depth of the neural network in the PSPNet we considered in experiments \cite{2017_PSPNet}. 

However, our approach has some limitations, that we point out hereafter, together with future works to be performed (or alternatives to be considered) to overcome them.
First, in the two  applications considered  in this paper, there is no effect of scale or rotation affecting the images so we did not seek invariance to such effects. So although our pipeline is invariant to translation, it is not  invariant to scaling effects (i.e. variation of distances between regions with respect to the model). Our approach could be extended to tackle scale effect  by using a relation invariant in scale (e.g. relative distance depending on the entire brain size instead of absolute spatial distance between brain regions) in the execution of the matching process. Concerning rotation, our approach is intrinsically invariant for the structural information regarding distances but not for the one based on relative positions (angles), particularly appropriate for the IBSR dataset. This limitation can be overcome by a preliminary detection of the overall orientation of the scene before applying the matching process (or it can be known from the image acquisition protocol, as is often the case in medical imaging). Note that the rotation invariance is also an issue of DNN-based segmentation, that is typically managed by data augmentation during training (i.e. applying rotations to training dataset).

Secondly, the first step (one-to-one) of the matching  procedure requires that each label is at least represented by a segmented region, which may not be the case in some particular situations (e.g. really small training dataset, as observed in the experiments with UNETR). Moreover, this first step is managed by only considering the largest probabilities from the segmentation map, ignoring lower one (although the entire probability map, therefore low probabilities, are taken into account for the refinement step). This could be overcome by taking into account lower probabilities at the early ``one-to-one'' step, or by managing, during the matching, situations where some vertices of the model are not present (e.g. ``one-to-one'' matching with a subgraph of the model graph). 

Thirdly, the fact that the application of our method mainly improves the Hausdorff distance and not the Dice coefficient is related to the segmentation provided by neural networks. Indeed, the segmentation contains classification errors that consist mainly of small regions far from the target region (artefacts). The fact that our method corrects to some extent these errors will then have a large impact on the Hausdorff distance, reflecting the important topological correction, but a small impact on the Dice coefficient (due to the small size of the corrected regions).

The last aspect to be discussed concerns the computation time, mainly affected by the refinement step involving many relabelling (illustrated in Figure \ref{fig:overview_ref}). In Algorithm \ref{algo_qap}, the complexity of this second step of the matching linearly depends on the cardinalities of both $U$ and $L$ entities as well as on the complexity of the cost computation (i.e. union of regions, $\text{Update-K}(R'_l)$, $vec(X)^T K vec(X)$ reported in lines 4-6 of Algorithm \ref{algo_qap}). Note that the first step may also involve a high computation time if all possible region candidates are considered (regardless their membership probabilities). In our case, one avoids this situation by initially trusting the DNNs output when considering only the most likely regions to be candidates to be matched with a given region of the model. Although trusting DNNs appears relevant in our experiments, one ignores candidates with a lower probability as previously mentioned. To consider all possible regions, the computation must be taken into account, and the complexity of the first one-to-one matching should be reduced, for instance by considering a progressive matching, starting by matching only few nodes of the model (subgraph matching), as recently studied with the use of a reinforcement-learning-based approach \cite{ICPRAI2022RL}.

\section{Conclusion}

We have proposed a post-processing technique for improving  segmentation results using a graph matching procedure encoding structural relationships between  regions. This correction of deep learning segmentation with the exploitation of structural patterns is performed thanks to  inexact graph matching formulated as a two-steps Quadratic Assignment Problem (QAP).
We validated our approach with experiments regarding two applications (2D image and 3D volumetric data) with various DNN backbones, and we have shown that significant improvements can be observed.    
When training the neural network on a limited dataset, our approach provides even more improvements.

Our proposal leads to a generic framework that can be extended to any kind of structural information (as illustrated with two types of relationships) and any other additional properties describing regions (as illustrated by combining both class membership probabilities and region size). Future work will focus on studying invariant representations to scale and orientation, on extending structural information to commonsense knowledge, and on applying our approach to additional applications.

\section*{Acknowledgments}
This research was conducted in the framework of the regional program Atlantic
2020, Research, Education and Innovation in Pays de la Loire, supported by the
French Region Pays de la Loire and the European Regional Development Fund.
This research was also partly conducted with the financial support of Science Foundation Ireland at ADAPT, the SFI Research Centre for AI-Driven Digital Content Technology at Maynooth University [13/RC/2106\_P2]. For the purpose of Open Access, the author has applied a CC BY public copyright licence to any Author Accepted Manuscript version arising from this submission.

\bibliographystyle{plain}
\bibliography{article_bib}

\clearpage

\appendix

\onecolumn

\section{FASSEG}
\label{sec:appendixFASSEG}

\begin{table}[ht!]
\caption{
Results obtained in the case of a U-Net neural network for different training configurations on the FASSEG dataset. The metrics used are the Dice score (DSC) and the Hausdorff distance (HD).}

\begin{tabular}{c|c|c|c|c|c|c|c|c|c|c|c|c}

\hline
Training & \multicolumn{4}{|c}{A training configuration} & \multicolumn{4}{|c}{B training configuration} & \multicolumn{4}{|c}{C training configuration}\\

\hline
Method & \multicolumn{2}{|c}{CNN} & \multicolumn{2}{|c}{Proposal} & \multicolumn{2}{|c}{CNN} & \multicolumn{2}{|c}{Proposal} & \multicolumn{2}{|c}{CNN} & \multicolumn{2}{|c}{Proposal}\\

\hline
Model & DSC$\uparrow$  & HD$\downarrow$  & DSC$\uparrow$  & HD$\downarrow$ & DSC$\uparrow$  & HD$\downarrow$ & DSC$\uparrow$  & HD$\downarrow$ & DSC$\uparrow$  & HD$\downarrow$ & DSC$\uparrow$  & HD$\downarrow$ \\

\hline
Hr & 0.90 & 117.01 & 0.90 & 115.62 & 0.90 & 115.22 & 0.90 & 120.34 & 0.88 & 123.89 & 0.88 & 122.01 \\
Fc & 0.90 & 37.57 & 0.90 & 35.32 & 0.91 & 38.62 & 0.91 & 29.53 & 0.90 & 51.98 & 0.90 & 38.86 \\
L-br & 0.58 & 36.60 & 0.58 & 20.75 & 0.60 & 31.34 & 0.60 & 23.48 & 0.59 & 34.67 & 0.60 & 25.70 \\
R-br & 0.53 & 27.80 & 0.53 & 32.75 & 0.58 & 29.44 & 0.58 & 24.50 & 0.60 & 53.49 & 0.61 & 36.00 \\
L-eye & 0.68 & 20.68 & 0.68 & 18.00 & 0.70 & 43.16 & 0.70 & 23.18 & 0.69 & 48.57 & 0.72 & 21.48 \\
R-eye & 0.66 & 23.15 & 0.66 & 31.10 & 0.71 & 16.84 & 0.71 & 16.33 & 0.72 & 41.92 & 0.73 & 16.40 \\
Nose & 0.61 & 17.56 & 0.61 & 16.34 & 0.65 & 39.75 & 0.67 & 13.22 & 0.62 & 24.06 & 0.62 & 18.43 \\
Mouth & 0.72 & 18.81 & 0.73 & 16.01 & 0.74 & 29.79 & 0.75 & 13.40 & 0.72 & 54.46 & 0.73 & 21.55 \\
\hline
avg & 0.70 & 37.40 & 0.70 & 35.74 & 0.72 & 43.02 & 0.73 & 33.00 & 0.71 & 54.13 & 0.72 & 37.56 \\
\hline
\end{tabular}
\end{table}

\begin{table}[ht!]
\caption{
Results obtained in the case of a U-Net neural network with a CRF on top for different training configurations on the FASSEG dataset. The metrics used are the Dice score (DSC) and the Hausdorff distance (HD).}

\begin{tabular}{c|c|c|c|c|c|c|c|c|c|c|c|c}

\hline
Training & \multicolumn{4}{|c}{A training configuration} & \multicolumn{4}{|c}{B training configuration} & \multicolumn{4}{|c}{C training configuration}\\

\hline
Method & \multicolumn{2}{|c}{CNN} & \multicolumn{2}{|c}{Proposal} & \multicolumn{2}{|c}{CNN} & \multicolumn{2}{|c}{Proposal} & \multicolumn{2}{|c}{CNN} & \multicolumn{2}{|c}{Proposal}\\

\hline
Model & DSC$\uparrow$  & HD$\downarrow$  & DSC$\uparrow$  & HD$\downarrow$ & DSC$\uparrow$  & HD$\downarrow$ & DSC$\uparrow$  & HD$\downarrow$ & DSC$\uparrow$  & HD$\downarrow$ & DSC$\uparrow$  & HD$\downarrow$ \\

\hline
Hr & 0.91 & 115.12 & 0.91 & 122.34 & 0.90 & 119.69 & 0.91 & 112.55 & 0.88 & 134.10 & 0.88 & 130.68 \\
Fc & 0.91 & 33.58 & 0.91 & 30.75 & 0.91 & 42.85 & 0.91 & 32.42 & 0.90 & 55.17 & 0.90 & 40.12 \\
L-br & 0.59 & 23.08 & 0.59 & 22.13 & 0.55 & 36.91 & 0.56 & 24.40 & 0.54 & 55.57 & 0.55 & 38.59 \\
R-br & 0.55 & 21.84 & 0.55 & 34.82 & 0.56 & 21.19 & 0.56 & 32.16 & 0.55 & 66.76 & 0.55 & 43.89 \\
L-eye & 0.68 & 20.86 & 0.68 & 17.01 & 0.70 & 25.19 & 0.70 & 21.17 & 0.66 & 55.44 & 0.68 & 39.53 \\
R-eye & 0.69 & 15.16 & 0.69 & 16.76 & 0.67 & 17.34 & 0.67 & 25.25 & 0.67 & 40.07 & 0.67 & 33.11 \\
Nose & 0.60 & 35.13 & 0.61 & 15.23 & 0.62 & 24.18 & 0.62 & 14.85 & 0.60 & 31.31 & 0.60 & 16.23 \\
Mouth & 0.74 & 18.38 & 0.75 & 14.93 & 0.72 & 31.64 & 0.73 & 17.78 & 0.72 & 41.64 & 0.72 & 25.54 \\
\hline
avg & 0.71 & 35.40 & 0.71 & 34.25 & 0.70 & 39.87 & 0.71 & 35.07 & 0.69 & 60.01 & 0.70 & 45.96 \\

\hline

\end{tabular}
\end{table}

\begin{table}[hp!]
\caption{
Results obtained in the case of a EfficientNet for different training configurations on the FASSEG dataset. The metrics used are the Dice score (DSC) and the Hausdorff distance (HD).}

\begin{tabular}{c|c|c|c|c|c|c|c|c|c|c|c|c}

\hline
Training & \multicolumn{4}{|c}{A training configuration} & \multicolumn{4}{|c}{B training configuration} & \multicolumn{4}{|c}{C training configuration}\\

\hline
Method & \multicolumn{2}{|c}{CNN} & \multicolumn{2}{|c}{Proposal} & \multicolumn{2}{|c}{CNN} & \multicolumn{2}{|c}{Proposal} & \multicolumn{2}{|c}{CNN} & \multicolumn{2}{|c}{Proposal}\\

\hline
Model & DSC$\uparrow$  & HD$\downarrow$  & DSC$\uparrow$  & HD$\downarrow$ & DSC$\uparrow$  & HD$\downarrow$ & DSC$\uparrow$  & HD$\downarrow$ & DSC$\uparrow$  & HD$\downarrow$ & DSC$\uparrow$  & HD$\downarrow$ \\

\hline
Hr & 0.93 & 83.96 & 0.93 & 85.31 & 0.93 & 95.70 & 0.93 & 93.73 & 0.91 & 109.36 & 0.91 & 108.28 \\
Fc & 0.95 & 23.42 & 0.95 & 19.23 & 0.95 & 29.51 & 0.95 & 21.28 & 0.94 & 48.48 & 0.95 & 27.53 \\
L-br & 0.73 & 13.32 & 0.73 & 10.63 & 0.73 & 27.35 & 0.73 & 11.99 & 0.69 & 38.51 & 0.69 & 20.76 \\
R-br & 0.73 & 14.75 & 0.72 & 14.49 & 0.71 & 14.40 & 0.72 & 17.87 & 0.69 & 51.94 & 0.69 & 18.66 \\
L-eye & 0.86 & 9.02 & 0.86 & 7.23 & 0.85 & 7.41 & 0.85 & 7.41 & 0.84 & 25.02 & 0.84 & 8.12 \\
R-eye & 0.83 & 14.12 & 0.83 & 11.27 & 0.84 & 15.44 & 0.84 & 11.13 & 0.82 & 70.75 & 0.84 & 12.52 \\
Nose & 0.80 & 9.04 & 0.80 & 7.23 & 0.79 & 10.46 & 0.79 & 7.48 & 0.76 & 27.99 & 0.75 & 12.33 \\
Mouth & 0.87 & 9.21 & 0.87 & 10.30 & 0.86 & 10.86 & 0.86 & 10.84 & 0.83 & 26.34 & 0.83 & 32.03 \\
\hline
avg & 0.84 & 22.11 & 0.84 & 20.71 & 0.83 & 26.39 & 0.83 & 22.72 & 0.81 & 49.80 & 0.81 & 30.03 \\

\hline

\end{tabular}
\end{table}

\begin{table}[hp!]
\caption{
Results obtained in the case of a PSPNet neural network for different training configurations on the FASSEG dataset. The metrics used are the Dice score (DSC) and the Hausdorff distance (HD).}

\begin{tabular}{c|c|c|c|c|c|c|c|c|c|c|c|c}

\hline
Training & \multicolumn{4}{|c}{A training configuration} & \multicolumn{4}{|c}{B training configuration} & \multicolumn{4}{|c}{C training configuration}\\

\hline
Method & \multicolumn{2}{|c}{CNN} & \multicolumn{2}{|c}{Proposal} & \multicolumn{2}{|c}{CNN} & \multicolumn{2}{|c}{Proposal} & \multicolumn{2}{|c}{CNN} & \multicolumn{2}{|c}{Proposal}\\

\hline
Model & DSC$\uparrow$  & HD$\downarrow$  & DSC$\uparrow$  & HD$\downarrow$ & DSC$\uparrow$  & HD$\downarrow$ & DSC$\uparrow$  & HD$\downarrow$ & DSC$\uparrow$  & HD$\downarrow$ & DSC$\uparrow$  & HD$\downarrow$ \\

\hline
Hr & 0.93 & 73.20 & 0.93 & 75.77 & 0.93 & 72.48 & 0.92 & 77.62 & 0.89 & 100.54 & 0.89 & 100.46 \\
Fc & 0.95 & 25.47 & 0.95 & 22.76 & 0.95 & 30.73 & 0.95 & 24.08 & 0.92 & 68.73 & 0.93 & 37.26 \\
L-br & 0.72 & 17.71 & 0.71 & 13.93 & 0.72 & 16.08 & 0.73 & 13.60 & 0.60 & 74.95 & 0.62 & 24.07 \\
R-br & 0.71 & 19.31 & 0.71 & 14.76 & 0.71 & 20.37 & 0.71 & 16.09 & 0.59 & 109.41 & 0.60 & 28.11 \\
L-eye & 0.85 & 11.80 & 0.85 & 7.42 & 0.85 & 11.03 & 0.85 & 8.14 & 0.73 & 46.16 & 0.72 & 20.76 \\
R-eye & 0.83 & 13.28 & 0.83 & 8.64 & 0.83 & 22.30 & 0.83 & 12.54 & 0.74 & 102.94 & 0.79 & 18.56 \\
Nose & 0.78 & 14.51 & 0.77 & 9.49 & 0.76 & 26.59 & 0.76 & 9.69 & 0.73 & 65.25 & 0.74 & 13.47 \\
Mouth & 0.86 & 18.14 & 0.86 & 12.87 & 0.85 & 22.93 & 0.85 & 9.34 & 0.81 & 84.09 & 0.80 & 23.49 \\
\hline
avg & 0.83 & 24.18 & 0.83 & 20.70 & 0.82 & 27.82 & 0.83 & 21.39 & 0.75 & 81.51 & 0.76 & 33.27 \\

\hline

\end{tabular}
\end{table}

\clearpage

\section{IBSR}
\label{sec:appendixIBSR}

\begin{table}[hp!]
\caption{
Results obtained in the case of a U-Net neural network for different training configurations on the IBSR dataset. The metrics used are the Dice score (DSC) and the Hausdorff distance (HD).}

\begin{tabular}{c|c|c|c|c|c|c|c|c|c|c|c|c}

\hline
Training & \multicolumn{4}{|c}{A training configuration} & \multicolumn{4}{|c}{B training configuration} & \multicolumn{4}{|c}{C training configuration}\\

\hline
Method & \multicolumn{2}{|c}{CNN} & \multicolumn{2}{|c}{Proposal} & \multicolumn{2}{|c}{CNN} & \multicolumn{2}{|c}{Proposal} & \multicolumn{2}{|c}{CNN} & \multicolumn{2}{|c}{Proposal}\\

\hline
Model & DSC$\uparrow$  & HD$\downarrow$  & DSC$\uparrow$  & HD$\downarrow$ & DSC$\uparrow$  & HD$\downarrow$ & DSC$\uparrow$  & HD$\downarrow$ & DSC$\uparrow$  & HD$\downarrow$ & DSC$\uparrow$  & HD$\downarrow$ \\

\hline

Tha.L & 0.82 & 54.18 & 0.90 & 5.45 & 0.83 & 47.87 & 0.90 & 5.70 & 0.81 & 46.80 & 0.90 & 5.66 \\
Tha.R & 0.86 & 45.52 & 0.90 & 4.18 & 0.83 & 45.65 & 0.90 & 3.94 & 0.83 & 45.20 & 0.90 & 4.23 \\
Cau.L & 0.86 & 18.36 & 0.86 & 4.23 & 0.87 & 17.82 & 0.87 & 3.97 & 0.83 & 16.79 & 0.83 & 5.66  \\
Cau.R & 0.85 & 7.16 & 0.85 & 4.74 & 0.85 & 7.81 & 0.85 & 4.87 & 0.83 & 15.67 & 0.83 & 5.04  \\
Put.L & 0.87 & 14.91 & 0.88 & 4.59 & 0.86 & 20.97 & 0.86 & 4.86 & 0.84 & 29.93 & 0.85 & 5.82  \\
Put.R & 0.88 & 6.85 & 0.88 & 4.29 & 0.88 & 14.14 & 0.88 & 4.64 & 0.85 & 19.38 & 0.86 & 4.94  \\
Pal.L & 0.82 & 3.93 & 0.82 & 3.93 & 0.83 & 5.74 & 0.83 & 3.63 & 0.80 & 5.51 & 0.80 & 3.80  \\
Pal.R & 0.78 & 6.84 & 0.78 & 4.09 & 0.78 & 9.53 & 0.78 & 4.07 & 0.73 & 13.65 & 0.73 & 4.38  \\
Hip.L & 0.79 & 65.38 & 0.82 & 5.38 & 0.73 & 58.52 & 0.81 & 5.74 & 0.74 & 65.14 & 0.79 & 6.87  \\
Hip.R & 0.80 & 49.52 & 0.82 & 7.30 & 0.79 & 54.91 & 0.82 & 7.31 & 0.77 & 52.20 & 0.80 & 7.08  \\
Amy.L & 0.76 & 11.59 & 0.76 & 4.58 & 0.73 & 10.57 & 0.73 & 4.77 & 0.72 & 25.03 & 0.72 & 4.95  \\
Amy.R & 0.76 & 21.90 & 0.76 & 4.79 & 0.74 & 20.53 & 0.75 & 5.07 & 0.70 & 33.09 & 0.71 & 5.45  \\
Acc.L & 0.68 & 4.96 & 0.68 & 3.81 & 0.68 & 3.67 & 0.68 & 3.67 & 0.62 & 4.24 & 0.62 & 4.24  \\
Acc.R & 0.68 & 3.99 & 0.68 & 3.99 & 0.64 & 4.16 & 0.64 & 4.16 & 0.58 & 4.57 & 0.58 & 4.57  \\
\hline
Avg & 0.80 & 22.51 & 0.81 & 4.67 & 0.79 & 22.99 & 0.81 & 4.74 & 0.76 & 26.94 & 0.78 & 5.19  \\
 
\hline

\end{tabular}
\end{table}

\begin{table}[h!]
\caption{
Results obtained in the case of a U-Net neural network with a CRF on top for different training configurations on the IBSR dataset. The metrics used are the Dice score (DSC) and the Hausdorff distance (HD).}

\begin{tabular}{c|c|c|c|c|c|c|c|c|c|c|c|c}

\hline
Training & \multicolumn{4}{|c}{A training configuration} & \multicolumn{4}{|c}{B training configuration} & \multicolumn{4}{|c}{C training configuration}\\

\hline
Method & \multicolumn{2}{|c}{CNN} & \multicolumn{2}{|c}{Proposal} & \multicolumn{2}{|c}{CNN} & \multicolumn{2}{|c}{Proposal} & \multicolumn{2}{|c}{CNN} & \multicolumn{2}{|c}{Proposal}\\

\hline
Model & DSC$\uparrow$  & HD$\downarrow$  & DSC$\uparrow$  & HD$\downarrow$ & DSC$\uparrow$  & HD$\downarrow$ & DSC$\uparrow$  & HD$\downarrow$ & DSC$\uparrow$  & HD$\downarrow$ & DSC$\uparrow$  & HD$\downarrow$ \\

\hline

Tha.L & 0.82 & 52.50 & 0.90 & 4.88 & 0.84 & 39.76 & 0.90 & 5.50 & 0.81 & 48.69 & 0.90 & 6.62 \\
Tha.R & 0.84 & 47.17 & 0.91 & 3.80 & 0.83 & 40.86 & 0.90 & 4.02 & 0.82 & 44.08 & 0.90 & 4.34 \\
Cau.L & 0.87 & 18.48 & 0.87 & 3.89 & 0.84 & 18.42 & 0.84 & 5.87 & 0.86 & 18.35 & 0.86 & 5.94 \\
Cau.R & 0.85 & 6.68 & 0.85 & 6.81 & 0.85 & 16.91 & 0.85 & 4.49 & 0.82 & 12.76 & 0.83 & 7.22 \\
Put.L & 0.89 & 25.08 & 0.89 & 4.32 & 0.88 & 13.74 & 0.88 & 4.59 & 0.86 & 34.97 & 0.87 & 4.85 \\
Put.R & 0.89 & 20.19 & 0.89 & 4.44 & 0.88 & 12.07 & 0.88 & 4.77 & 0.87 & 13.53 & 0.87 & 4.86 \\
Pal.L & 0.84 & 4.84 & 0.84 & 3.52 & 0.82 & 7.24 & 0.82 & 3.72 & 0.81 & 12.36 & 0.82 & 3.72 \\
Pal.R & 0.78 & 4.05 & 0.78 & 4.05 & 0.77 & 6.93 & 0.77 & 4.28 & 0.76 & 17.44 & 0.77 & 3.76 \\
Hip.L & 0.78 & 65.50 & 0.82 & 4.95 & 0.74 & 64.75 & 0.82 & 5.48 & 0.72 & 67.68 & 0.81 & 5.94 \\
Hip.R & 0.81 & 51.96 & 0.84 & 7.06 & 0.79 & 64.27 & 0.83 & 5.72 & 0.78 & 52.87 & 0.82 & 6.78 \\
Amy.L & 0.74 & 35.35 & 0.75 & 4.75 & 0.73 & 14.69 & 0.73 & 4.94 & 0.72 & 39.98 & 0.74 & 4.87 \\
Amy.R & 0.76 & 18.82 & 0.76 & 4.75 & 0.76 & 21.31 & 0.76 & 4.93 & 0.72 & 27.73 & 0.72 & 5.22 \\
Acc.L & 0.73 & 6.28 & 0.73 & 3.50 & 0.67 & 3.72 & 0.67 & 3.72 & 0.65 & 4.13 & 0.65 & 4.13 \\
Acc.R & 0.69 & 4.63 & 0.69 & 3.83 & 0.63 & 4.18 & 0.63 & 4.18 & 0.59 & 4.45 & 0.59 & 4.79 \\
\hline
Avg & 0.81 & 25.82 & 0.82 & 4.61 & 0.79 & 23.49 & 0.81 & 4.73 & 0.77 & 28.50 & 0.80 & 5.22 \\

\hline

\end{tabular}
\end{table}

\begin{table}[h!]
\caption{
Results obtained in the case of a UNETR for different training configurations on the IBSR dataset. The metrics used are the Dice score (DSC) and the Hausdorff distance (HD).}

\begin{tabular}{c|c|c|c|c|c|c|c|c|c|c|c|c}

\hline
Training & \multicolumn{4}{|c}{A training configuration} & \multicolumn{4}{|c}{B training configuration} & \multicolumn{4}{|c}{C training configuration}\\

\hline
Method & \multicolumn{2}{|c}{CNN} & \multicolumn{2}{|c}{Proposal} & \multicolumn{2}{|c}{CNN} & \multicolumn{2}{|c}{Proposal} & \multicolumn{2}{|c}{CNN} & \multicolumn{2}{|c}{Proposal}\\

\hline
Model & DSC$\uparrow$  & HD$\downarrow$  & DSC$\uparrow$  & HD$\downarrow$ & DSC$\uparrow$  & HD$\downarrow$ & DSC$\uparrow$  & HD$\downarrow$ & DSC$\uparrow$  & HD$\downarrow$ & DSC$\uparrow$  & HD$\downarrow$ \\

\hline
Tha.L & 0.85 & 11.50 & 0.85 & 5.63 & 0.84 & 15.25 & 0.84 & 6.32 & 0.82 & 34.25 & 0.83 & 14.25 \\
Tha.R & 0.85 & 17.25 & 0.85 & 4.78 & 0.84 & 22.41 & 0.84 & 4.59 & 0.79 & 31.67 & 0.79 & 8.31 \\
Cau.L & 0.79 & 15.07 & 0.79 & 6.35 & 0.73 & 29.67 & 0.74 & 5.90 & 0.62 & 29.82 & 0.62 & 6.71 \\
Cau.R & 0.78 & 19.75 & 0.78 & 7.16 & 0.76 & 6.19 & 0.76 & 5.17 & 0.60 & 38.80 & 0.61 & 6.36 \\
Put.L & 0.84 & 17.61 & 0.84 & 4.91 & 0.82 & 10.22 & 0.82 & 4.93 & 0.74 & 35.92 & 0.76 & 5.47 \\
Put.R & 0.85 & 13.59 & 0.85 & 4.63 & 0.82 & 26.50 & 0.82 & 5.05 & 0.74 & 44.33 & 0.77 & 4.86 \\
Pal.L & 0.77 & 8.39 & 0.77 & 4.15 & 0.75 & 12.63 & 0.75 & 4.85 & 0.69 & 21.51 & 0.70 & 4.74 \\
Pal.R & 0.77 & 6.48 & 0.77 & 3.39 & 0.75 & 6.17 & 0.75 & 4.38 & 0.71 & 28.79 & 0.72 & 3.81 \\
Hip.L & 0.69 & 16.56 & 0.69 & 8.65 & 0.65 & 13.63 & 0.65 & 8.75 & 0.62 & 23.35 & 0.63 & 11.73 \\
Hip.R & 0.67 & 9.79 & 0.67 & 7.77 & 0.66 & 10.68 & 0.66 & 9.62 & 0.60 & 35.71 & 0.59 & 12.31 \\
Amy.L & 0.66 & 5.35 & 0.66 & 5.35 & 0.66 & 15.36 & 0.67 & 5.26 & 0.70 & 16.19 & 0.70 & 4.87 \\
Amy.R & 0.66 & 5.33 & 0.66 & 5.33 & 0.64 & 13.36 & 0.64 & 5.62 & 0.63 & 5.49 & 0.63 & 5.49 \\
Acc.L & 0.69 & 3.85 & 0.69 & 3.85 & 0.65 & 7.02 & 0.66 & 3.22 & 0.51 & 10.09 & 0.51 & 5.60 \\
Acc.R & 0.68 & 3.61 & 0.68 & 3.61 & 0.64 & 3.62 & 0.64 & 3.62 & 0.53 & 5.26 & 0.53 & 5.33 \\
\hline
Avg & 0.75 & 11.01 & 0.75 & 5.40 & 0.73 & 13.76 & 0.73 & 5.52 & 0.67 & 25.80 & 0.67 & 7.13 \\

\hline

\end{tabular}
\end{table}

\end{document}